\documentclass{ws-rv975x65}
\usepackage{subfigure}     
\usepackage{ws-rv-thm}     
\usepackage{ws-rv-van}     
\makeindex

\begin{document}

\chapter[Robot Autonomy for Surgery]{Robot Autonomy for Surgery}\label{ra_ch1}

\author[M. Yip and N. Das]{Michael Yip and Nikhil Das}

\address{University of California, San Diego,\\
9500 Gilman Drive, La Jolla, CA, 92093\\
\{yip, nrdas\}@ucsd.edu}

\begin{abstract}
Autonomous surgery involves having surgical tasks performed by a robot operating under its own will, with partial or no human involvement. There are several important advantages of automation in surgery, which include increasing precision of care due to sub-millimeter robot control, real-time utilization of biosignals for interventional care, improvements to surgical efficiency and execution, and computer-aided guidance under various medical imaging and sensing modalities. While these methods may displace some tasks of surgical teams and individual surgeons, they also present new capabilities in interventions that are too difficult or go beyond the skills of a human. In this chapter, we provide an overview of robot autonomy in commercial use and in research, and present some of the challenges faced in developing autonomous surgical robots.
\end{abstract}
\body

\section{Introduction}
Robot autonomy refers to the performance of tasks where part or all of the task is completed by an intelligent robotic system. Since the use of robotics in surgery in the mid-1980s, applying autonomy to surgery has been a continuing effort for engineers and researchers. As with conventional human-performed surgery, surgeries performed autonomously through robotic systems require these systems to be safe and precise in their movements, to adequately sense the environment and the patient, and to make decisions and adapt to changes in the environment under different situations. There is a large spectrum of autonomous behaviors that can be observed in surgical procedures: some are far easier to implement and encourage the use of (such as tremor reduction), while others are much more forward-looking, such as autonomous cardiac ablation in the beating heart, where a clinician must rely on the robot entirely to creating continuous lesions in the heart. The technologies associated with these works are often a combination of multiple areas of research and technology: robot design and control, medical image integration and real-time signal processing, and artificial intelligence and machine learning.

In this chapter, we will consider the following topics in order:
\begin{enumerate} 
\item define and distinguish the levels of autonomy for surgical robotics,
\item consider the potential benefits of autonomy in surgery,
\item describe systems for surgical autonomy in commercial use,
\item describe research efforts in hardware and software toward surgical automation,
\item detail current challenges in research and development for automated surgery, and
\item discuss ethical and legal concerns involved in automating surgery.
\end{enumerate}
We conclude with future directions for robot autonomy in surgery. 

While we will highlight specific technologies and developments of interest and importance to surgical autonomy, this chapter is not meant to provide a complete report on all efforts on surgical autonomy, since an exhaustive search would bring up thousands of examples including efforts that may or may not be considered automation; instead, we will use the topics above to frame the topic of surgical autonomy and explore key strategies, innovations, and considerations in its implementation.

\section{Scope of Automation}
\begin{figure}[b]
\centerline{\includegraphics[width = \textwidth]{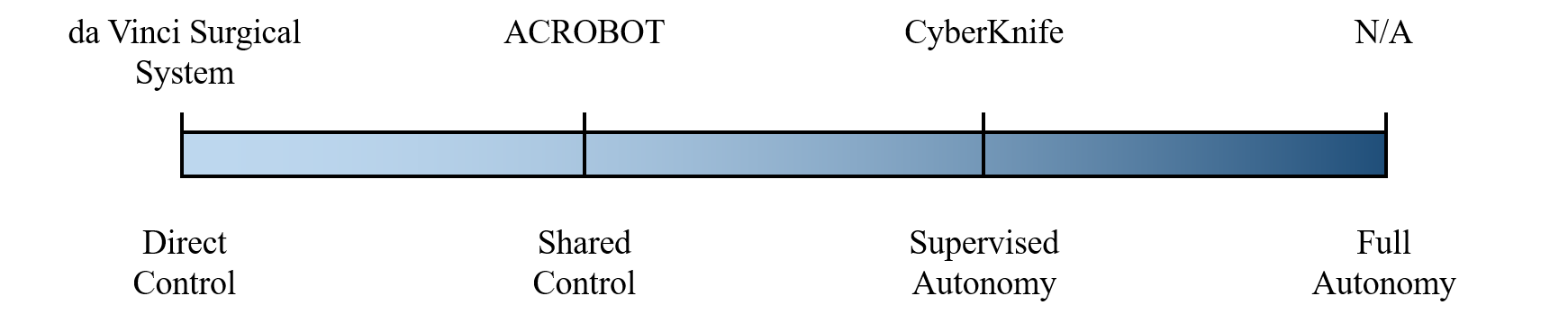}}
\caption{Spectrum of autonomy in robotic surgery and examples of systems. There are currently no examples of surgical systems that can operate without any level of human involvement.}
\label{spectrum}
\end{figure}

	The spectrum of automation in surgical robotics relates to the level of dependence on the human surgeon to guide the procedure, as compared to the robot guiding itself. These levels are \textit{direct control}, \textit{shared control}, \textit{supervised autonomy}, and \textit{full autonomy}. Fig. \ref{spectrum} illustrates this range of autonomy and provides examples of commercial systems at each level. Here we define the different automation methods at a high level, though many different commercial systems may fall within two or more categories.
    
    In \textit{direct control}, human surgeons are in complete control of the surgical tasks by either manually controlling or teleoperating the robotic manipulators. In a teleoperation scheme, the surgeon sits at a user control console with joysticks and a camera feed, including access to any other pertinent biosignals or imaging modalities placed in the same display, to remotely operate the tools. The robot directly obeys the human surgeon's commands and therefore does not move by itself; however, one can consider some form of robot automation involved in the mapping of complex robot joints velocities into tool velocities, a complex mapping that is masked from the human operator. This level of autonomy can be found in most teleoperated systems. The da Vinci Surgical System, developed by Intuitive Surgical Inc. in 2000 was the first FDA-approved robotic system for laparoscopic surgery and is an example of such a system. The surgical instruments working on the patient are teleoperated by a human surgeon sitting at a console in a non-sterile section of the operating room\cite{Talamini2002}. Indeed, the da Vinci system is capable of fully autonomous behavior as we will see in future sections, but is currently not approved for autonomous motions.

The next level of autonomy is \textit{shared control}. As the name suggests, shared control is where different types of surgical tool motions are simultaneously distributed in their control between the human and robot\cite{Moustris2011,Kang2001}. The Steady Hand robot, developed at Johns Hopkins University, is a system designed for sub-millimeter manipulation tasks such as retinal microsurgery\cite{Taylor1999}. The Steady Hand system allows the human surgeon to control all movements of the surgical instrument but gradually applies a counteractive force proportional to the force sensed by the tool tip\cite{Kapoor2003}. The net motion of the robot manipulator is thus a based on a combination of the human's movements and the robot's application of force feedback to correct for tremor, making the Steady Hand an example of a shared control system.

At the intermediate level is \textit{supervised autonomy}, where the robot executes surgical tasks with some autonomy but under supervision of a human surgeon. The human oversees the operation and issues high-level commands while the robot executes the motions. This enables the surgeon to still remain in charge of the decision-making process, but with less involvement in the execution. The CyberKnife system, invented at Stanford University and now built by Accuray\cite{Moustris2011}, generates and executes preoperative plans to deliver radiotherapy to tumors located on the patient's body. The human surgeon adjusts the automatically generated plan prior to execution and ensures the system performs the task safely during the procedure\cite{Tombropoulos2011}.

    At the end of the spectrum is \textit{full autonomy}, where a robotic system completely supplants the intraoperative role of the surgeon. Such a system will autonomously plan and execute entire surgical operations. As surgical procedures are complex and intricacies of the operation depend on patient-specific information, full autonomy in surgery is currently not likely to be realized in practice in any near future. However, it is a potential goal in automated healthcare\cite{Sharkey2013} in that it could significantly improve the consistency and quality of care to the patient population, as well as be distributed to remote areas for improving patient access to high-quality surgical care. 
    
    While full autonomy of entire surgical procedures is currently far-fetched even from a technological point of view, entire procedures may be decomposed into smaller surgical tasks, such as suturing or resection, that are more feasible to automate. Section \ref{approachesToAutomation} discusses recent approaches to surgical automation of certain surgical tasks in research. Similar to human language, surgical activities can be described systematically in terms of the constituent parts, i.e., the basic motions that the surgeons perform with their instruments or hands\cite{Gao2014}. Thus, decomposition of surgical procedures may be taken a step further by identifying the primitive surgically-relevant motions (called surgemes) that make up the execution of a given task\cite{Gao2014}. 

\section{Advantages of Autonomy in Surgery}
There are many advantages to automating parts or all of a surgery. Human surgeons, while great at making high-level decisions regarding the care of the patient, are less capable at performing precision tasks, especially in the face of fatigue and under hand tremors. This causes the variability of fine motor tasks and, by extension, surgical care, to be large between experts and novices, and from metropolitan areas to remote communities. With many of these surgeons overworked to high levels of fatigue, they are less attentive and more prone to human error. Autonomous surgery provides a consistency and quality to a treatment unaffected by these issues. 

A second major advantage of automating robots for performing surgery is that a robotic system can provide significantly greater dexterity in its tools than a human-controlled tool\cite{Siciliano2008}. Increasing dexterity of tools has been shown to be effective in minimally-invasive procedures. Such advantages can already be observed in current robotic systems such as the da Vinci Surgical System, which affords greater dexterity during laparoscopic surgery through its wristed manipulators\cite{Heemskerk2006,Kang2001a}. Additionally, endoscopic robots and catheters, including the Hansen Medical Sensei System\cite{Gomes2011}, allow the clinician to position the tip of a catheter without concerning her or himself with the manipulation of a catheter handle. This avoids learning the complex mappings from the proximal handle to the distal tip of catheter, which can take years to master\cite{Chun2008}. In these systems, another compelling benefit is that the surgeon no longer needs to be in the same room and thus can avoid stray radiation from X-ray fluoroscopy\cite{Chun2008,Davies2000}. 

Medical imaging, including X-ray fluoroscopy, MRI, CT, and ultrasound, provide important subsurface and volumetric information of a patient's anatomy and can be used both preoperatively to plan or intraoperatively to guide a procedure\cite{boogerd2016}. The potential benefits from incorporating medical imaging as real-time feedback for a surgical procedure cannot be understated; its use in automating the biopsy and delivery of therapy has been shown to improve surgical precision and accuracy, reduce margins, avoid sensitive tissues and nerves, and improve consistency of treatments\cite{Muntener2006,Alterovitz2008,Vardo2017,Abayazid2014}.

\begin{table}[t]
\tbl{Advantages and disadvantages of humans and autonomous robotic systems}
{\begin{tabular}{@{}ccc@{}} \toprule
 & Human & Robot \\
\colrule
Advantages & Good judgement & Good mechanical precision \\
& Adaptable and able to improvise & Untiring and stable \\
& Able to use qualitative information & Can work in hazardous environments\\
& Easy to train & Multimodal sensory integration\\
& Easy communication with humans & \\

\colrule
Disadvantages & Limited mechanical precision & No judgement  \\
& Prone to fatigue, tremor, inattention  & No qualitative abilities\\
 & Cannot work in hazardous environments & Limited in haptic sensation \\
 & Limited quantitative abilities & Expensive \\

\botrule
\end{tabular}
}
\label{humanVsRobotTable}
\end{table}

While robotic systems benefit humans in terms of mechanical ability, a great deal of research is still needed before they can provide judgment and planning capabilities that approach human cognition. Preoperative planning is better suited for human surgeons. Automation in current surgical robotic systems thus potentially provides advantages including the ability to execute predefined motions in complex environments accurately and tirelessly, to restrict paths or positions of instruments to satisfy safety constraints, or to respond quickly to changing environments based on sensors or commands\cite{Davies2000}. Systems such as the CyberKnife and ACROBOT demonstrate some of these advantages. Table \ref{humanVsRobotTable} summarizes advantages and disadvantages of the capabilities of human surgeons and automated robotic surgical systems.

Autonomy in surgery may also benefit administratively by increasing hospital throughput. There is a greater consumption of skilled surgical staff during a teleoperated robotic surgery with the da Vinci Surgical System compared with traditional laparoscopic surgery\cite{Link2006,Lanfranco2003}. During a robotic surgery, the team involved in the intraoperative procedures includes bedside assistants standing near the patient who perform assistive and often routine tasks including tissue retraction, staple application, suction, and irrigation. Since the bedside assistant must perform these tasks manually, the assistant does not experience the same benefits of the primary surgeon sitting at a surgical console where 4D stereoscopic and egocentric displays\cite{Ballantyne2002} and tremor filtering\cite{Heemskerk2006,Kang2001a} may be available. Not only does this degrade the bedside assistants' capabilities and places them in potentially awkward ergonomics, it also results in poor fine-motor coordination between them and the surgeons. Automation of the routine assistive tasks alleviates the bedside assistants from the disadvantages they face and could allow redistribution of these skilled surgical staff to other areas of the hospital. 

\begin{table}[t]
\tbl{Robotic systems with various levels of autonomy}
{\begin{tabular}{@{}ccc@{}} \toprule
Name & Branch of surgery & Level of Autonomy \\
\colrule
da Vinci Surgical System & General minimally-invasive & Direct control \\
EndoBot & General minimally-invasive & Direct, shared control, supervised \\
Trauma Pod & General & Direct control \\
Sensei Robotic System & Cardiac & Direct control \\
NeuroMate & Neurosurgery & Supervised \\
Probot & Urologic & Supervised \\
ACROBOT & Orthopedic & Shared control \\
RIO & Orthopedic & Shared control \\
Precision Freehand Sculptor & Orthopedic & Shared control \\
ROBODOC & Orthopedic & Supervised \\
CyberKnife & Radiosurgery & Supervised \\

\botrule
\end{tabular}
}
\label{systemListTable}
\end{table}

\section{Application Areas}
Surgical automation can provide several significant benefits over manual surgeries, from increased precision and accuracy, improved consistency in treatments, greater dexterity and access to tissues, more effective utilization of medical imaging, etc. Despite these many benefits, the adoption of autonomy in the surgical domain is still at its infancy. 

Surgical procedures can be considered in three automation-relevant stages\cite{Manzey2009}:
\begin{enumerate}
\item{information analysis and acquisition}
\item{plan generation, and }
\item{execution of surgical action.}
\end{enumerate}
The information analysis and acquisition stage may include preoperative imaging such as CT or MRI scans and the identification of supplementary landmarks. Automation of this stage may use image processing and pattern recognition algorithms to assist in the location of unique features that may be used as reference points\cite{Manzey2009}. The plan generation stage involves making decisions based on the information gathered in the first stage. Automated systems may suggest trajectories or insertion points to the surgeon during this stage. The execution stage is when the robot carries out the trajectories either autonomously or with partial human assistance. To be able to carry out a surgical plan autonomously, the system must be able to correlate the landmarks identified in the first stage to the patient in the operating room\cite{Eggers2006}. Procedures for which each of these stages may be automated are suitable candidates to incorporate autonomy in clinical practice. This section describes examples in different branches and types of surgery for which various surgical stages are automated.

\subsection{Orthopedic Surgery}
One of the most important considerations for autonomy is to match a precomputed treatment plan based on preoperative medical images with anatomical features on the patient in the operating room. Between the preoperative and intraoperative stages, patient anatomies may shift, and they may continue to shift over the course of the operation. Thus, it not surprising that orthopedics was among the first areas to utilize automation because bones can generally be considered non-deformable and are thus easier to manipulate\cite{Howe1999}. Autonomous execution of orthopedic surgical tasks, such as milling predefined areas in bone as a preparatory step for implant surgery, is consequently more accurate and consistent to preoperative plans than work with soft tissues.

The ROBODOC system, sold by THINK Surgical Inc.\cite{Netravali2016}, is used for Total Knee Arthroplasty procedures, which requires the formation of cavities in the bone prior to placing the prosthetic implant. Prior to the ROBODOC cavity formation, titanium locator pins are implanted in the patient's leg which are used as fiducial markers or supplementary landmarks for the information acquisition (via CT scan) stage\cite{Spencer1996}. Based on these CT scans and fiducial markers, the software system generates motion plans that the surgeon verifies and supervises during execution\cite{Spencer1996}. Clinical trials demonstrate that compared to manually performing the procedure, automated robotic cavity sculpting with the ROBODOC leaved a smoother surface which is more consistent with the preoperative plan\cite{Spencer1996}. Systems like the ROBODOC are often referred to as surgical CAD/CAM systems, alluding to high-precision computer-aided design (CAD) and manufacturing (CAM) of a part, which are analogous to the planning stage and the execution stage respectively\cite{Kazanzides2008}.

\begin{figure}[ht]
\centerline{\includegraphics[width = 1.3in]{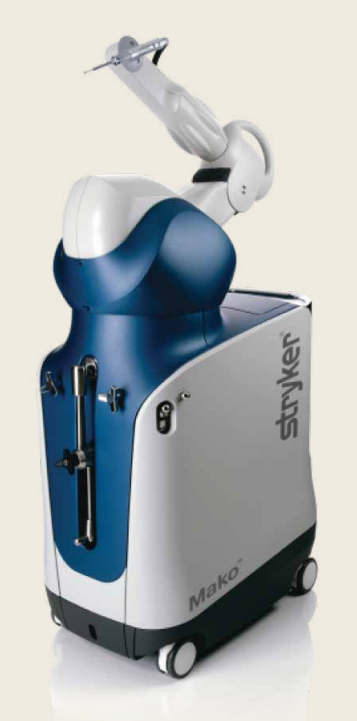}}
\caption{MAKO/Stryker Surgical System\cite{VanderList2016} is an example of a shared control system, as the robot applies constrains to the surgeon's actions when approaching the boundaries of the safe region.}
\label{mako}
\end{figure}

Another recent example of automation in othopedic surgery includes the MAKO/Stryker Surgical System\cite{Hagag2011} (Fig. \ref{mako}), which operates on a different principle than the ROBODOC. While the ROBODOC acts nearly like a milling machine on an autonomous route, the MAKO System acts as a hand-held robotic arm, that knows internally the cutting boundaries of the bone but do not drive the milling; instead, the surgeon drives the milling tool around until the tool reaches a \textit{virtual fixture}, virtual structures that constrain the surgeon's movements to a certain volume\cite{Rosenberg1993}. These virtual fixtures serve as a guide to the clinician. There is little technological reason as to why the surgeon needs to hold the robot tool, but it more comes down to the FDA approval process and how much of the surgical procedure was deemed safe to hand over to a computer agent (see Section \ref{approvalForUse}). Regardless of the technology, the basic principles of these technologies rest on the fact that bone is stiff and nondeformable, and thus preoperative plans can be registered without fail and without loss of accuracy to intraoperative anatomy.


\subsection{Neurosurgery}
Automation has also been applied to minimally-invasive neurosurgery for operations such as deep brain stimulation and stereotactic electroencephalography\cite{Moustris2011}. As is the case with bones in orthopedic surgery, the skull provides a rigid frame of reference during surgery. Unfortunately, brain tissue is soft and requires intraoperative imaging to account for changes in the tissue during the operation.

Frameless stereotaxy, the precise localization of target sites inside a patient using 3D coordinates without fixing the patient to a rigid frame\cite{Barnett2007}, is possible with the NeuroMate system by Renishaw using an ultrasound probe to locate fiducial markers placed on the patient's head which are compared with CT or MRI scan images\cite{Li2002,Varma2006}. The planning software locates the markers and plans the trajectory based on the high-resolution preoperative images\cite{Varma2006}. During the actual operation, an ultrasound transmitter and receiver system captures the same points as located during the preoperative imaging and planning stage\cite{Li2002}. Triangulation of the markers is performed intraoperatively using a time-of-flight method, in which a signal processing unit estimates distances by measuring the time delay between each ultrasound transmitter-receiver pair\cite{Li2002}. If the same procedure were performed using a frame-based method, i.e. the patient's head is mechanically fixed in place, this ultrasound-based intraoperative localization step is ignored as fixation of the patient's head suggests the preoperative images are still aligned with the current position of the patient\cite{Li2002,Varma2006}. The NeuroMate is thus able to use the preoperative imaging to generate a plan and position itself for frameless stereotaxy in preparation for further procedures such as deep brain stimulator implantation. The NeuroMate ultimately can not only lessen the need for a human surgical tool holder assistant but also provide safer and more cost-efficient assistance\cite{Li2002}. Trials demonstrate that the frameless stereotaxy method does not perform as well as frame-based stereotaxy, but still is able to achieve an error of less than 2 mm for 3D tracking\cite{Li2002}.

\subsection{Radiosurgery}
Autonomous execution of motion plans is also used in stereotactic radiosurgery, a procedure that uses focused beam of radiation for tumor treatment\cite{Tombropoulos2011}. Contactless radiotherapy systems typically fire multiple intersecting beams of radiation from various angles to maximize radiation dosage to the target site while minimizing radiation to healthy tissue, known as isocentric targeting\cite{Dieterich2011}.

The CyberKnife system differs from other radiotherapy systems like the Gamma Knife\cite{Lindquist1995} by generating motion plans for nonisocentric targeting, which promotes homogeneous dosage over irregular targets\cite{Tombropoulos2011,Dieterich2011}. During the information analysis and acquisition stage, a surgeon defines the regions of interest from CT or MRI scans. The motion planner named CARABEAMER represents these regions of interest as 3D volumes and generates a plan to maximize dosage the target regions\cite{Tombropoulos2011}. During the execution of the surgical plan, the patient is not mechanically fixed to a frame. The CyberKnife system provides frameless treatment using image-guided stereotactic targeting\cite{Barnett2007}. Using periodic X-ray images, the CyberKnife tracks patient movement using fiducial markers. Unlike the NeuroMate, however, the CyberKnife does not require implanted fiducial markers, but instead uses external fiducial markers to track patient movement such as a snugly-fit vest with optical markers to track motion due to breathing. The internal target locations are updated in real-time based on a correlation function to the external markers, located using the X-ray images\cite{Sayeh2007}. The CyberKnife system ultimately is able to provide a contactless radiotherapy treatment\cite{B2004} based on an automatically-generated surgical plan and real-time updating through image-guided targeting.

\subsection{Brachytherapy}
Automation for image-guided procedures is useful for precise tasks such as needle steering for brachytherapy source placement\cite{Hungr2009,Cunha2010}. Brachytherapy is a procedure used for cancer treatment involving placing radioactive sources, or seeds, near tumors\cite{Muntener2006}. Aligned with the three automation-relevant stages of surgery, the clinical workflow for brachytherapy treatment involves image acquisition of the patient, radiation dose and seed placement planning, and seed placement via needles\cite{Cunha2010}. 

These stages have been automated in an experimental environment using the MRBot, a MRI-compatible system developed at Johns Hopkins University\cite{Stoianovici2011}, for prostate brachytherapy\cite{Cunha2010}. Image acquisition may be performed using either MRI or CT imaging modalities. A special marker is installed on the tip of the robot to facilitate in aligning the frame of the world to the frame of the images. Planning software analyzes the images by identifying organ locations and determining needle patterns, seed placement locations, and radiation dosages. MRI guidance was used to place needles through tissue phantoms and non-homogeneous bovine tissue and to verify the needle placement prior to seed delivery\cite{Cunha2010, Muntener2006}. The average error for needle placement in tissue phantoms is 0.72 mm, demonstrating the precision of the planning software and robotic system\cite{Muntener2006}.

Current research seeks to improve image-guided needle steering to account for uncertainties due to patient differences and non-homogeneous tissues\cite{Alterovitz2008}. This approach uses a Markov decision process, a popular model for random systems in which a robot must make decisions, to maximize the probability the needle reaches the goal location. 

\subsection{Telesurgery}
Telesurgery refers to the use of robotic systems by a human surgeon to assist a geographically-separated patient\cite{Haidegger2011} and is useful for remote or dangerous environments such as the battlefield, space, and underserved regions of the world\cite{Siciliano2008}. When the patient is in a location inaccessible to the medically trained humans, automation of surgically-relevant tasks is crucial.

The Trauma Pod system, founded by the US military-technology agency DARPA\cite{Romaniuk2017}, is a deployable robotic system that is a contained surgical suite with a da Vinci-like system, robotically controlled imaging, and a robot assistant, and is designed for the treatment of soldiers in critical condition and out of range of a combat hospital\cite{Garcia2009}. Without the availability of human assistants, the Trauma Pod must autonomously handle all primary and assistive tasks, including preoperative image acquisition and intraoperative tool changes\cite{Garcia2009}. The entire Trauma Pod system is divided into subsystems which facilitate the automation of the surgical stages. The Patient Registration Subsystem (PRS) generates a 3D model of the patient which is used to safely maneuver the manipulators around the patient\cite{Rosen2011}. The Patient Imaging Subsystem handles the image acquisition of the patient through X-ray fluoroscopy device that is moved in a grid pattern above the patient\cite{Rosen2011,Garcia2009}. The Resource Monitoring Subsystem (RMS) records intraoperative clinically-relevant events and protocols, such as supply usage, fluids, and event times of surgical procedures\cite{Garcia2009}. The actual surgical operation is not performed autonomously, but instead a remote surgeon interacts with the Trauma Pod through the User Interface Subsystem at a surgical master console, which may also provide a 3D visualization of a simulated reconstruction of the operating room generated with the Simulator Subsystem (SIM)\cite{Garcia2009}. The SIM is capable of high-resolution collision detection and real-time model updating at a 30 Hz rate\cite{Garcia2009}. Via verbal commands, the human surgeon commands the Scrub Nurse Subsystem (SNS) to handle subsystem calibration, deliver supplies, and exchange surgical tools\cite{Garcia2009}, tasks similar to a human scrub nurse. Thus, while the Trauma Pod does not perform the surgical intervention itself, the Trauma Pod successfully automates the three stages of surgical procedures as a bedside assistant for tool changing during telesurgeries. The PRS handles the information-gathering stage for the operating environment. The Supervisory Control Subsystem serves as the task planner, managing the high-level control of the subsystems and scheduling the execution of all assistive events\cite{Rosen2011}. The SNS, PIS, and RMS execute the assistive tasks such as tool swapping and patient monitoring\cite{Garcia2009}.

\subsection{Intracardiac Surgery}
For research purposes, intraoperative surgical tasks are often considered in isolation, i.e. preoperative information-gathering and planning stages are less emphasized. Instead, more emphasis is placed on technology for intraoperative care, being able to adapt to the environment during the execution stage or to autonomously perform the intraoperative task with more accuracy or speed than manual performance. For example, intracardiac surgery involves procedures in the heart, a challenging environment due to the narrow space and high-signal disturbances due to heartbeats. Cardiac ablation procedures, a common treatment for arrhythmia, requires scarring tissue to correct abnormal heartbeats. Manual control of the ablation catheters is difficult for even well-trained surgeons, and thus control of such catheters is a suitable candidate for automation.

\begin{figure}[b]
\centerline{
  \subfigure[]
     {\includegraphics[height=1.7in]{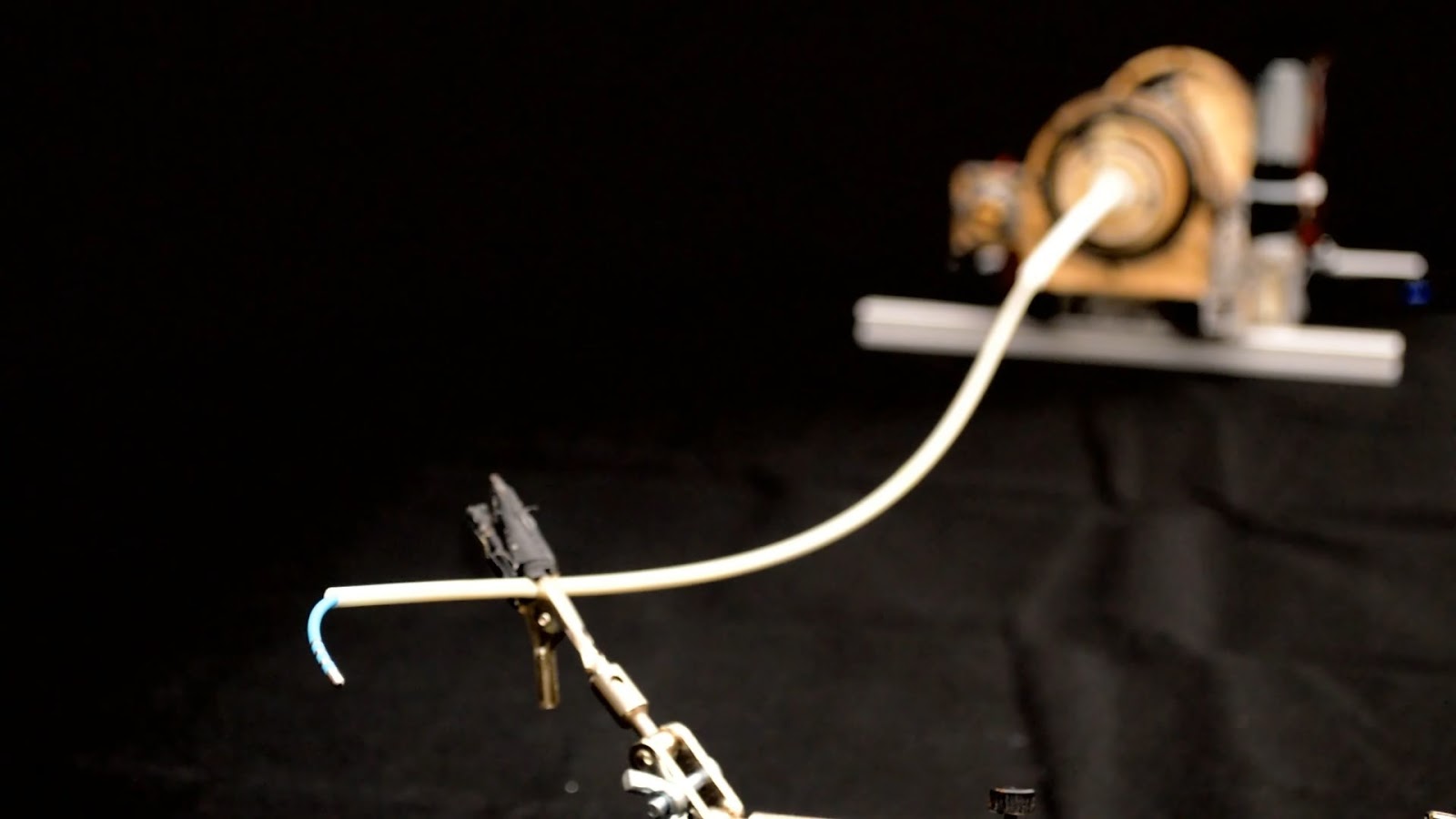}\label{continuum}}
  \hspace*{4pt}
  \subfigure[]
     {\includegraphics[height=1.7in]{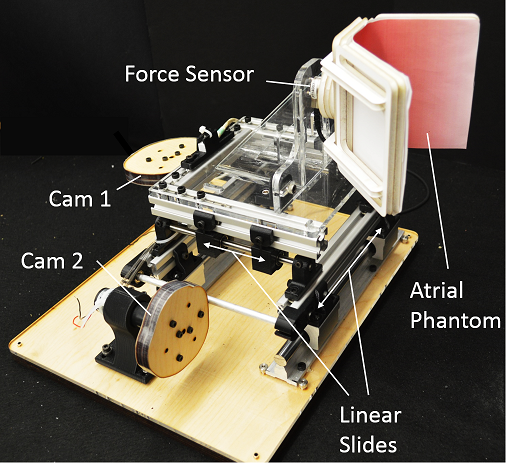}\label{CuPID}}
}
\caption{(a) Continuum robot system for cardiac ablation and (b) Simulated beating heart environment for testing path-tracing tasks\cite{Yip2017}. Automating treatments using robotic catheters and endoscopes are still in their infancy, challenged by the complexities of robot control and limited sensing.}\label{continuum}
\end{figure}

The task of autonomous control for catheters robots (called continuum robots due to their continuously flexible body) has been studied for use in cardiac ablation procedures\cite{Yip2017}  (Figure \ref{continuum}). Because maintaining direct contact with the tissue surfaces is critical to treatment success, a robot is used to trace a heart wall with constant applied force in the presence of beating disturbances. The challenge with these types of robots (both catheter robots and endoscopic robots) is that that the control mappings from proximal to distal ends change arbitrarily due to how the devices are constrained, thus making them unsuitable for traditional physics model-based robot controllers to work with. A viable strategy is to learn on-the-fly how to control these devices by estimating the robot Jacobian mappings in real-time which may be used to determine how much to change an actuator for a desired change in the tool position. Compared with manual control of the continuum robot, the automated control can successfully traced desired paths by dragging the catheter around the heart\cite{Yip2017}, a task too challenging for human to control through a manual catheter (they typically use point-by-point ablations) . This is an example of where robot controlled devices can present a new, potentially more effective method of treatment. Ultrasound image guidance has also been proposed using intracardiac echography, can can be controlled in an autonomous fashion to keep important features in view\cite{loschak2013}, while others have considered fluoroscopic image guidance for guiding catheters in vasculature\cite{jayender2008}. However, these works are still in the research phase, and most continuum robotic systems in use for surgery are directly teleoperated\cite{burgner2015}.

\section{Autonomy in Commercially-Available Surgical Systems}
Some degree of autonomy in the form of shared control and supervised autonomy is available on commercial surgical systems. The current state of autonomy employed on these commercial systems requires a high level of preoperative or intraoperative human involvement, and the tasks which are automated are typically tasks that can be routinized. These commercial systems have been successfully implemented in clinical practices because they combine the computational and mechanical advantages of robotic systems with the contextual and high-level intelligence of a human surgeon.

\begin{figure}[b]
\centerline{
  \subfigure[]
     {\includegraphics[height=3.4in]{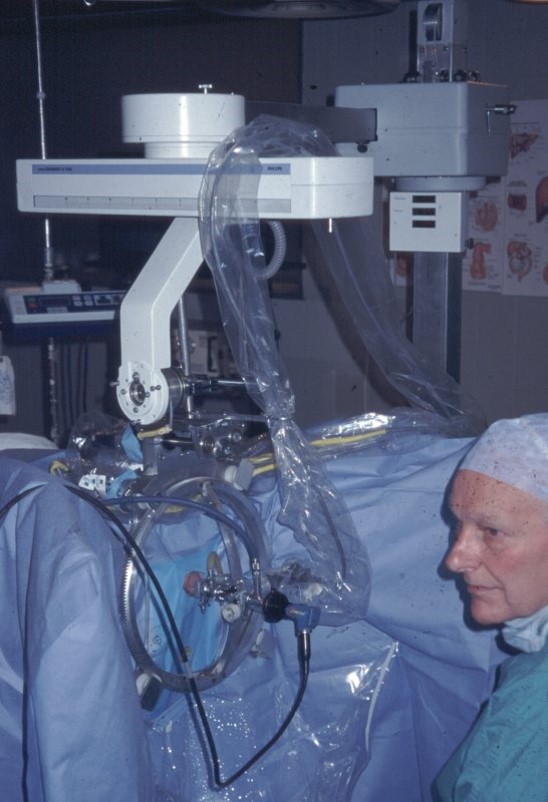}\label{probot}}
  \hspace*{4pt}
  \subfigure[]
     {\includegraphics[height=3.4in]{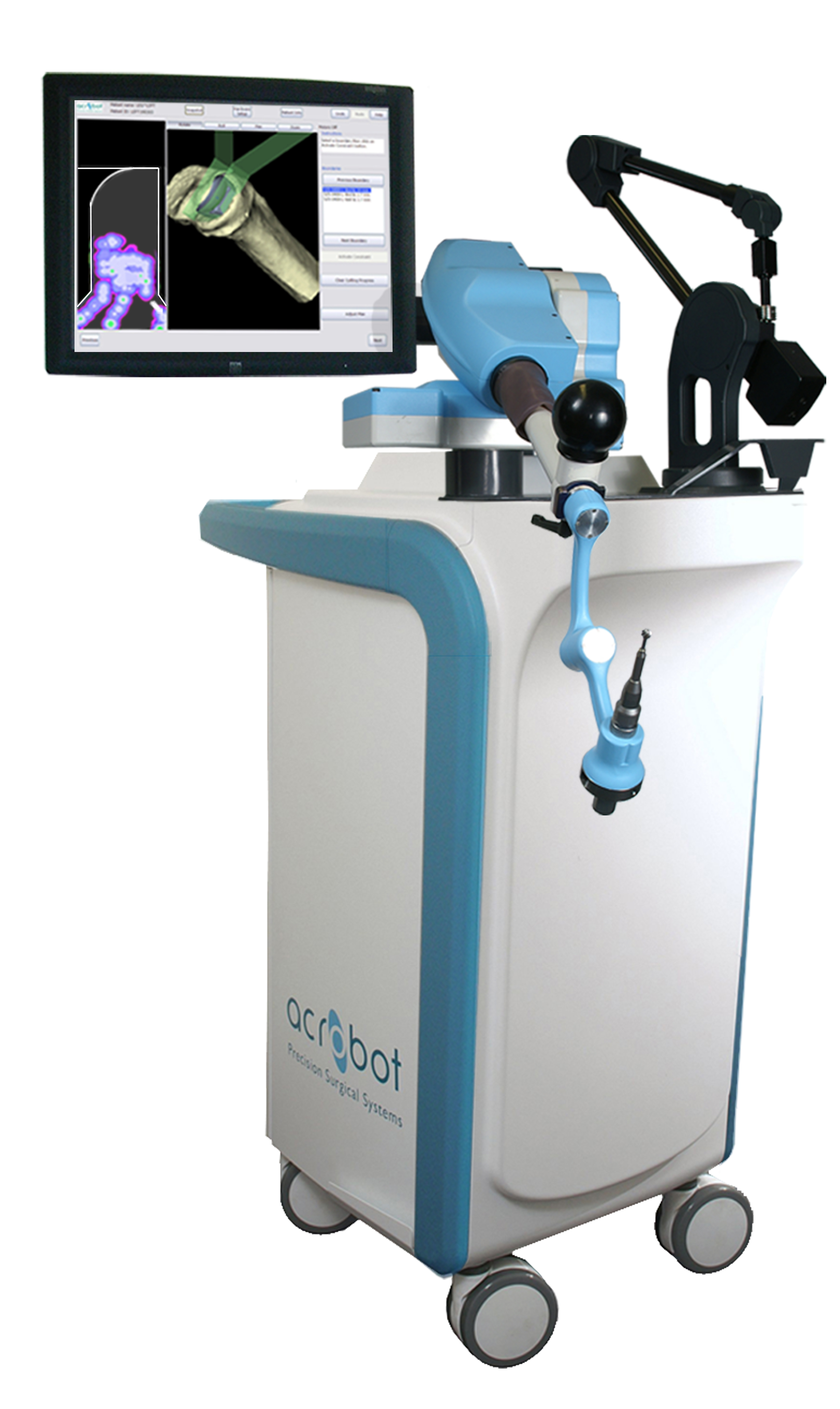}\label{acrobot}}
}
\caption{(a) Probot system, a supervised autonomous system which carries out autonomous prostate resection motions. (b) ACROBOT system, a shared control system which constrains surgeon-controlled movements of the surgical tool to a safe region. Reproduced by permission of Dr. Brian Davies, Imperial College London.}\label{probotAcrobot}
\end{figure}

The Probot, developed in 1991 at Imperial College London, is one of the first instances of supervised autonomy in surgery (Fig. \ref{probot}). The robot was used for transurethral prostate resection. After appropriate positioning of the system by a human, the Probot could autonomously remove conical segments of tissue while the human surgeon controlled an emergency stop switch\cite{Harris1997,Mei1996}. Playing a passive role as the robot performed actions autonomously caused uneasiness for the surgeons, so a system in which the surgeon plays a more active role in the operation was desired.\cite{RodriguezyBaena2010}

The ACROBOT (Active Constraint Robot), which began development in 1991 at Imperial College and later marketed by the Acrobot Company Ltd.\cite{RodriguezyBaena2010}, is one of the first examples of shared autonomy in robotic surgery (Fig. \ref{acrobot}). Unlike the Probot, the human surgeon is in control of the ACROBOT during operation. The ACROBOT is designed for precise bone cutting prior to knee replacement surgery. The ACROBOT is used as an intelligent surgeon-controlled tool that constrains movements within a predefined safe region\cite{RodriguezyBaena2010}. If the surgeon attempts to move out of the safe region, the surgeon feels some resistance in the controller prohibiting further erroneous motions\cite{Jakopec2002}, similar to virtual fixtures deployed on the Mako Surgical system by Stryker \cite{Hagag2011}. 

The ROBODOC and CyberKnife systems are examples of supervised autonomy, which are used for orthopedic surgery and stereotactic radiosurgery, respectively. Both systems utilize preoperative imaging to create motion plans and can execute preoperative plans without human interruption excluding emergency situations. A difference between the ROBODOC and CyberKnife is in how the patient registration problem is addressed. Patient registration is the process of relating preoperative images to the position of the patient\cite{Eggers2006}, which is a crucial step when executing preoperative image-based plans on the actual patient. The ROBODOC system requires the patient to be rigidly held in place and preoperative images are utilized to generate and execute surgical plans. A position monitor discontinues the operation if the patient leg moves\cite{Rosen2011}. The CyberKnife system uses real-time imaging and adjustment to track the position of the patient to account for movement, such as that due to respiration\cite{Moustris2011,Sayeh2007}.

\section{Research Platforms for Surgical Automation}
Automation in surgery is a developing field, and thus an active area of research. Open source platforms exist to facilitate research in automation in surgical robotics. These platforms are useful to develop and test new automation techniques in simulated or artificial environments before progressing to in vivo clinical trials.

\begin{figure}
\centerline{\includegraphics[width = \textwidth]{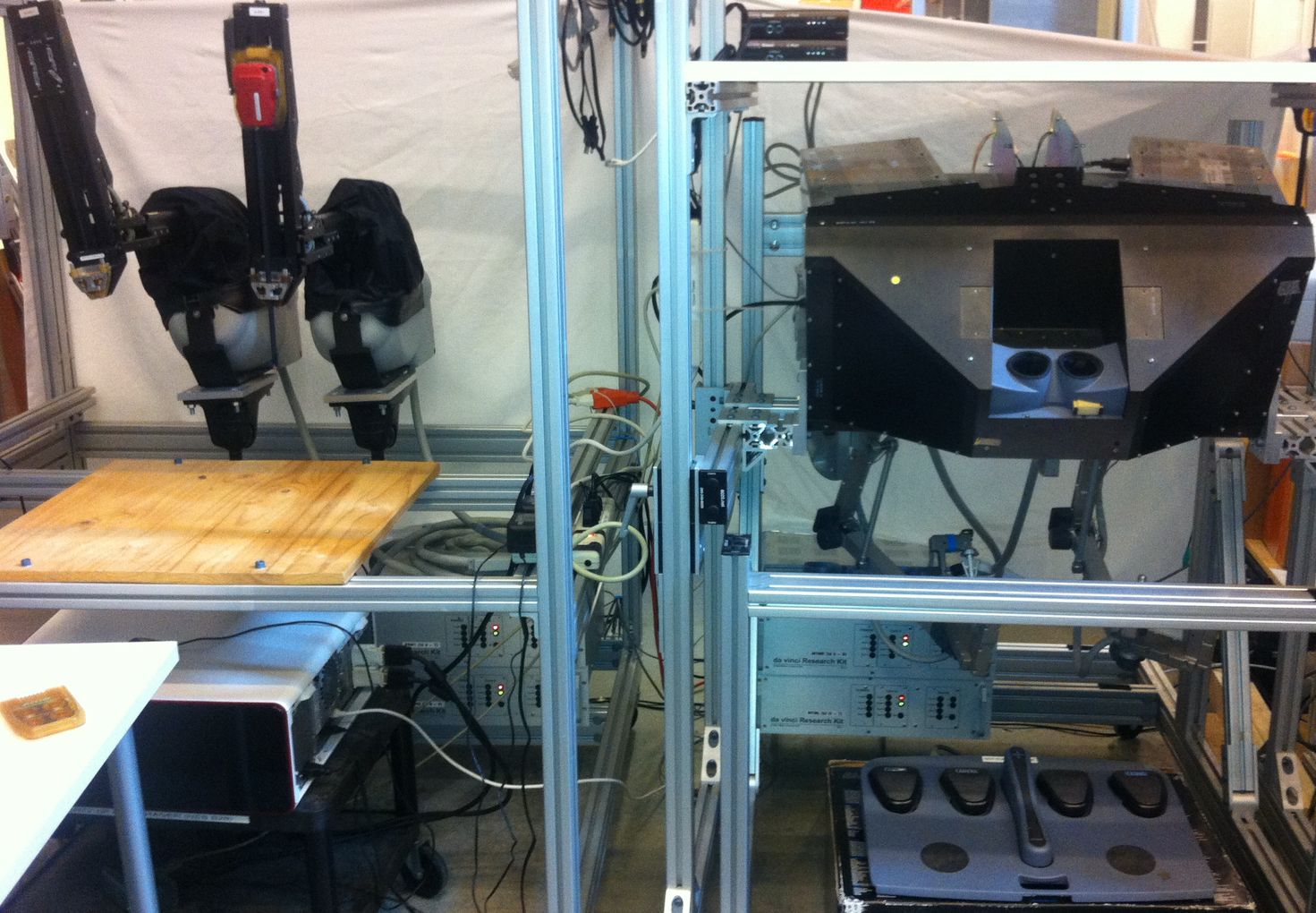}}
\caption{The da Vinci Research Kit provides a stripped down, open-access hardware and software interface for researching autonomy in surgical scenarios. The major advantage of this system is that the hardware matches the clinical model and thus provides a standardized platform for development as well as an easy facilitator towards translation into future da Vinci systems.}
\label{dvrk}
\end{figure}

The da Vinci Research Kit (dVRK) is based on the first generation da Vinci Surgical System and enables control-level access to the system\cite{Chen2013}. Either a retired first-generation da Vinci system or a subset of the components provided by Intuitive Surgical, Inc., may be used for the hardware of the research kit. The subset includes two Master Tool Manipulators, two Patient Side Manipulators (PSMs), a High Resolution Stereo Viewer, and a footpedal tray. The full system additionally includes a third PSM, an Endoscopic Camera Manipulator, and passive Setup Joints\cite{Chen2017} (Fig. \ref{dvrk}. All computation is centralized on a high-performance PC while multiple FPGAs handle the I/O communication. Centralizing computation and distributing I/O enable flexible reconfigurations of the system, such as converting a bimanual teleoperation system into two independent unilateral systems\cite{Chen2013}. The da Vinci System uses components of the Surgical Assistant Workstation (SAW) software package for low-level I/O, joint-level and high-level control, and teleoperation\cite{Chen2013}. SAW attempts to standardize the interfaces with preexisting open-source standards such as OpenIGTLink used for image-guided therapy\cite{Kazanzides2010}. Current research has used the dVRK for debridement, pattern cutting\cite{Murali2015}, and ultrasound transducer placement\cite{Mohareri2014} tasks. 

\begin{figure}[hb]
\centerline{\includegraphics[width = \textwidth]{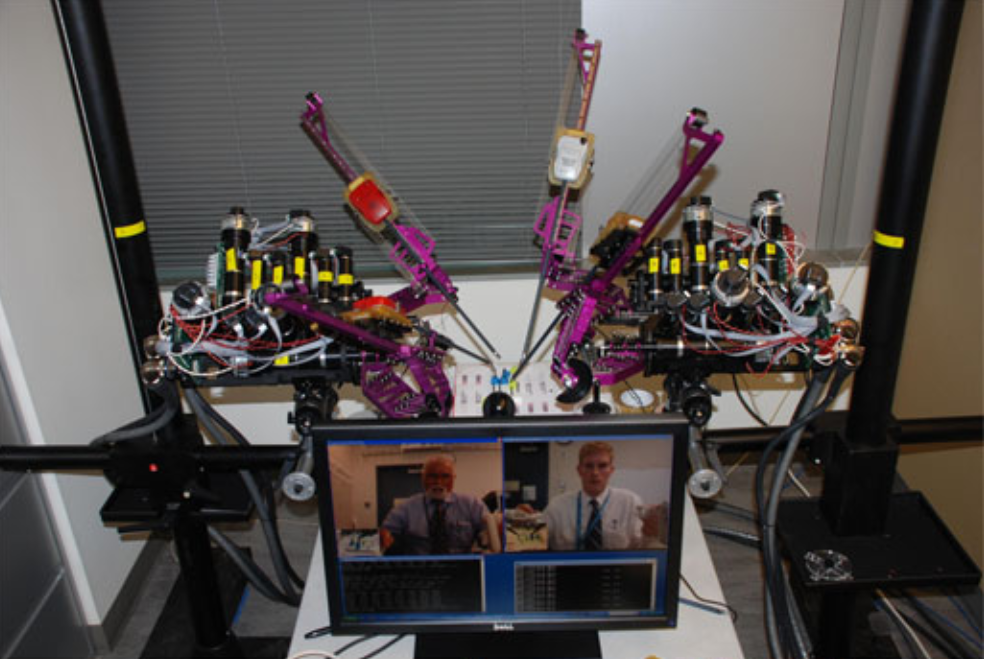}}
\caption{Two surgeons in University of Washington collaborate on a surgical task using a Raven II surgical system located at the University of California, Santa Cruz\cite{Hannaford2013}.}
\label{raven}
\end{figure}

The Raven II, developed at the University of Washington and currently managed by Applied Dexterity, Inc, is an open-architecture system for laparoscopic surgery and is designed to facilitate research in surgery (Fig. \ref{raven}). Compared to the dVRK, the Raven II system is designed to encourage collaboration among a network of researchers\cite{Hannaford2013}. This system includes two cable-driven laparoscopic arms with wristed graspers. Unlike the da Vinci system, the Raven II is designed to be portable and small, enabling two Raven II's to be mounted on both sides of the surgery site allowing a total of four arms in the same space. Having two separate systems on the same surgical site enables two independent surgeons to collaborate on the same surgery. The users teleoperate the system through the PHANTOM Omni haptic device controller. The surgeon interfaces with the system over the internet using a TCP/UDP protocol\cite{Hannaford2013} enabling the user to operate the system from anywhere on the globe. The Raven II system has been used to automate tasks such as tumor ablation\cite{Hu2015} and debridement\cite{Kehoe2014}.

\begin{figure}[t]
\centerline{\includegraphics[width=\textwidth]{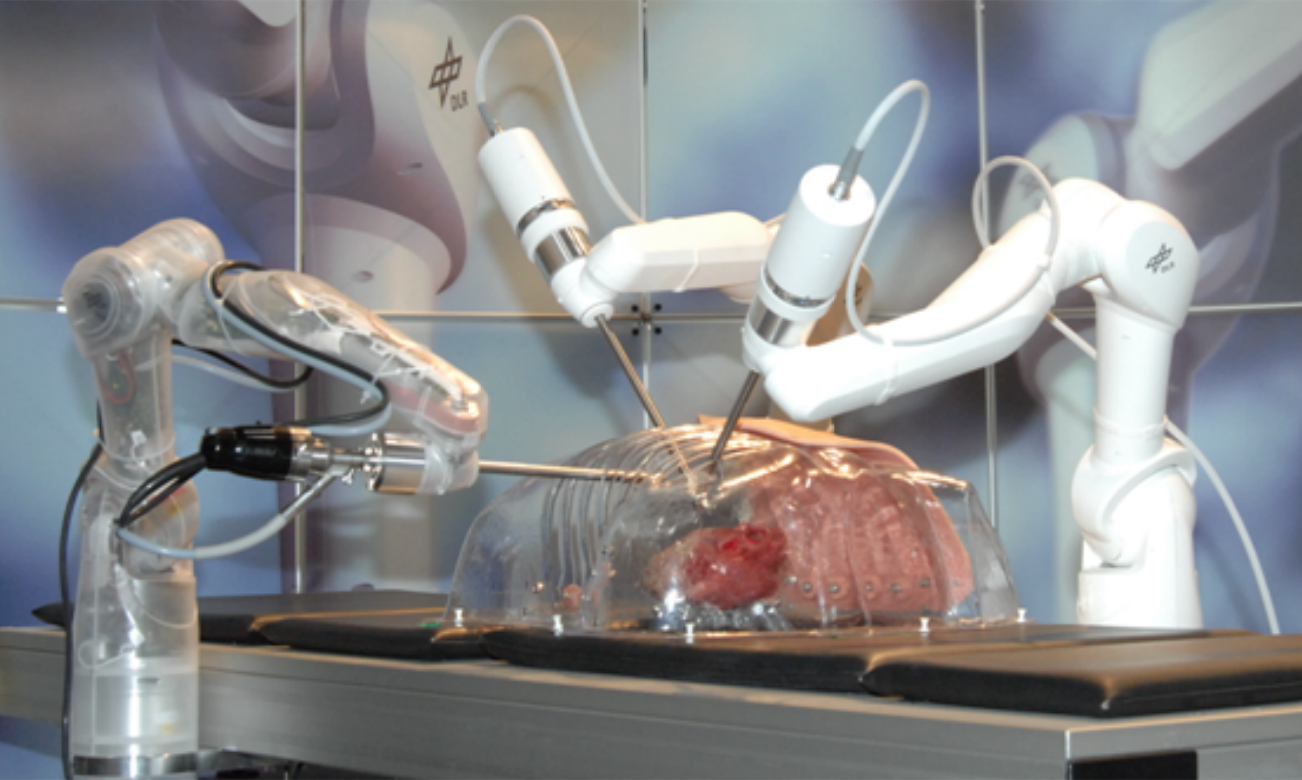}}
\caption{MiroSurge system developed at DLR with table-mounted manipulators\cite{Hagn2010}.}
\label{mirosurge}
\end{figure}

The MiroSurge system is in development at the German Aerospace Center specifically for use as a research platform for multiple endoscopic surgical applications\cite{Hagn2010} (Fig. \ref{mirosurge}). The majority of the system's components, such as position sensors and motor brakes, are on the manipulators, which reduces the external components. The robot manipulators are lightweight and allow ceiling or wall mounting, enabling adaptability to the operating room. Unlike the da Vinci system which has a physically fixed remote-center-of-motion for each manipulator, the MiroSurge manipulators use null space constraints to artificially enforce fulcrum point constraints. Null space constraints enable changes to the joint angles of the manipulator with certain points of the arm fixed in space. The advantage of using null space constraints is the MiroSurge has the ability to arbitrarily define the fulcrum point, thereby increasing the system's versatility. While capable of autonomous surgery, this system has seen little use outside of DLR's research team, though its capabilities surpass those present in the daVinci system and the Raven II. 

One of the key ingredients to the continued development and success of surgical autonomy on robotic platforms is the open-sourced nature of the research platforms and the common testbeds of use. Both the daVinci and the Raven II has been used by over a dozen institutions worldwide, who provide full schematics to the hardware and software for their base systems through online, version-controlled repositories. Another approach is to use highly-engineered industrial robotic arms as a tool for exploring automation without concerning oneself with the open-source maintained and buggy nature of the above systems. The Kuka Lightweight Robot Arm, one of the most popular industrial robot arms in the world, has been used successfully for surgical automation in microanastamosis tasks. While not cleared for surgical tasks, the affordance of a industrial-robot-precision arm offers great advantages for researchers interested in the higher-level autonomy, as demonstrated in Shademan et al\cite{Shademan2016}.

\section{Approaches to Automation}
\label{approachesToAutomation}
Various techniques have been employed to automate surgical tasks. The techniques described in this section involve a human in a preplanning stage, utilize control theory to follow a human during the operation, or use machine learning techniques to learn behaviors or motions from human-provided examples. Many of the techniques described here are also highlighted in other sections as features such as visual servoing or haptic constraints can be found in both research and commercial systems.

\subsection{Predefining Motions or Constraints}
\begin{figure}[t]
\centerline{
  \subfigure[]
     {\includegraphics[height=2in]{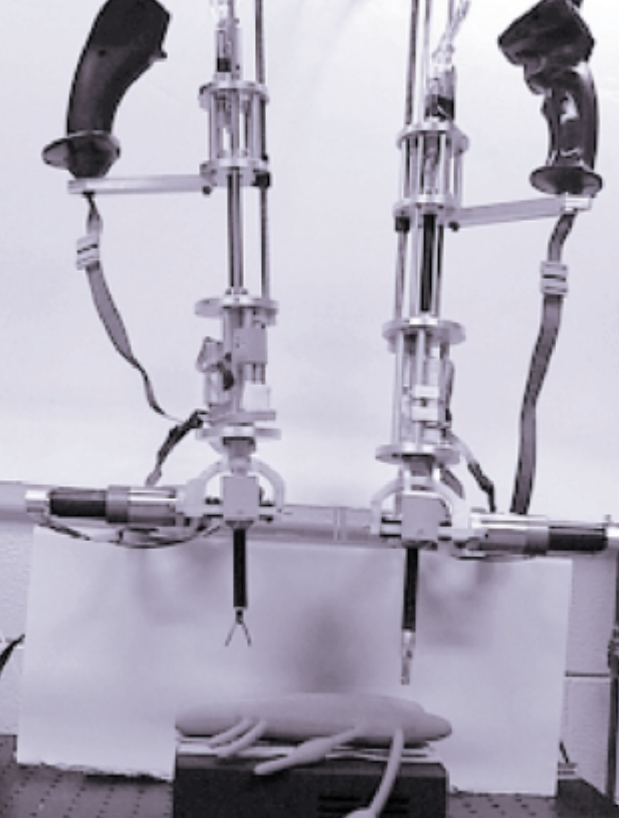}\label{endobot1}}
  \hspace*{4pt}
  \subfigure[]
     {\includegraphics[height=2in]{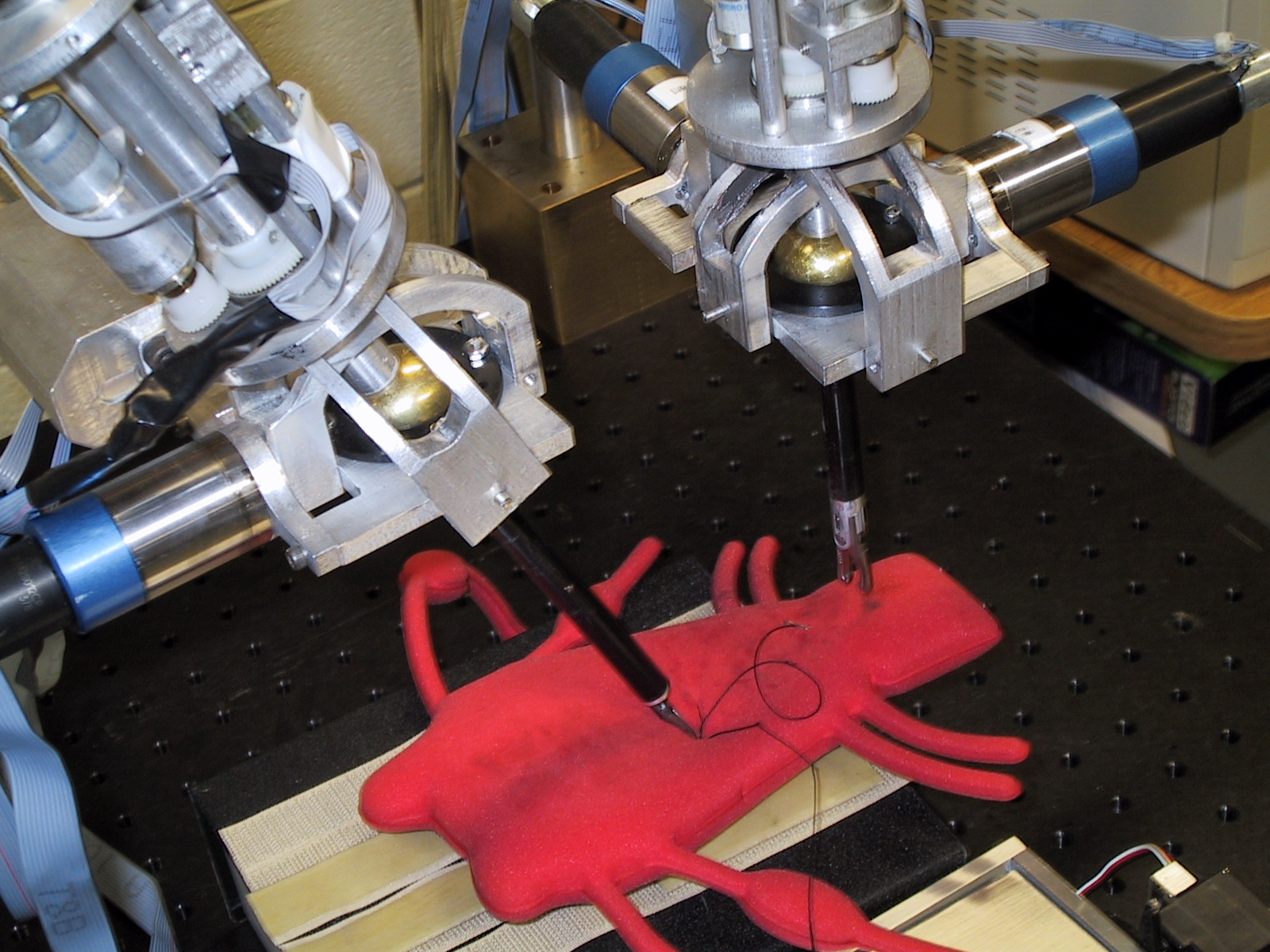}\label{endobot2}}
}
\caption{(a) EndoBot system. (b) Automated suturing performed using EndoBot system on simulated tissue. Reproduced by permission of Dr. John Wen, Rensselaer Polytechnic Institute.}\label{endobot}
\end{figure}

The simplest form of surgical automation is predefining and executing a sequence of steps without updating the motion plan in response to a different environment. Minimally-invasive suturing was accomplished by executing a predefined and fixed sequence of steps using the EndoBot (Fig. \ref{endobot}), a system designed to allow direct, shared, and supervised autonomous control of the manipulators\cite{Kang2001,Kang2001a}. While this method may work for controlled environments and tasks, execution of steps without a method to adapt to the environment will generally fail in real surgical situations if there is movement. Nevertheless, defining a fixed sequence of steps may be beneficial as an initial exemplary motion model for future improvements.


The ROBODOC system uses a software system called ORTHODOC to assist in planning the motions required for femoral milling\cite{Rosen2011}. Titanium locator pins are first implanted in the patient's leg which are used as fiducial markers\cite{Spencer1996}. Preoperative CT images are uploaded to the ORTHODOC system which creates a surgical plan which uses the locator pins for patient registration, i.e., the process of determining the correspondence between medical images and the position of the actual patient. A human surgeon verifies and manipulates the plan generated by the ORTHODOC. During the actual operation, the surgeon fixes the femur to the robot base to limit movement and moves the robotic arm to the locator pin locations to complete the patient registration process. The ROBODOC system then performs the milling procedure based on the predefined motion plan under the surgeon's supervision\cite{Spencer1996}.

While predefining motions directs the robot where to go, predefining constraints tells the robot where not to go. A common approach to prevent movement into restricted regions is through the use of virtual fixtures\cite{Rosenberg1993}. The ACROBOT, used for orthopedic surgery, constrains movements of the cutting tool to a predefined region of the operative field\cite{Rosen2011}. The ACROBOT does not move autonomously, but instead the surgeon manually controls the instruments. The ACROBOT imposes the workspace constraints via haptic feedback when the surgeon attempts to move the tools outside the safe region. The boundary constraints are determined preoperatively from CT scans of the patient using an interactive planning software developed at Imperial College. The software generates a 3D model of the bones, and based on the surgeon-specified prosthesis size and location, the software defines the constraints\cite{Jakopec2002}.

The Robotic Arm Interactive Orthopedic (RIO) System, developed by MAKO Surgical Corporation, also utilizes the idea of virtual fixtures. Similar to the ACROBOT, the RIO system provides haptic feedback on the controller when the surgeon attempts to move outside the predefined safe volume.\cite{Hagag2011}. However, unlike the ACROBOT, the RIO system can keep track of changes in tissues and update the virtual model of the patient's knee based on fiducial markers\cite{Rosen2011}. This flexibility to account for changes even enables intraoperative revisions of preoperative plans\cite{Hagag2011}.

\subsection{Visual Tracking and Servoing}

Visual servoing defines the use of images in a feedback loop to automate the robotic system. Thus, a significant number of efforts in automating surgery are intrinsically directly related to or an extension of visual servoing problems. One task to highlight is the positioning of tools relative to underlying soft tissues, which is difficult to do as it requires both tracking of the anatomy and maintaining registration to the underlying anatomy\cite{yip2012}. Thus, integration of real-time imaging, particularly ultrasound, provides a useful means for visualizing tissue locations \cite{Vitrani2005}. Often, it can be difficult to discern tissues from instruments, and thus ridged instrument markers may be used to facilitate locating the instrument in 3D space, where the ridges assist in identifying the instrument's position and orientation with a single ultrasound image\cite{Stoll2006}. 

Another area that has garnered considerable investigation in visual servoing for surgery is autonomous endoscopic camera movement. Since endoscopic cameras for minimally invasive surgery are usually manually moved by an assistant or the primary surgeon, this task of endoscopic control is a low-reaching goal. Surgeons typically position the camera such that the area being worked on is in clear or panoptic view. One study has shown autonomously adjusting the camera can reduce the number of camera corrections per hour by a factor of approximately 7 when using an autonomous camera control system versus manual control\cite{Omote1999}. Autonomous endoscopes typically move according to the surgeon's intent, but inferring the surgeon's intent is a challenge in itself. One method to measure the motions that may indicate the surgeon's intent is through instrument tracking, which works under the assumption that the location of the tools indicate the location of the surgeon's attention\cite{Pandya2014}. The AESOP system, a robotic endoscope developed by Computer Motion typically controlled by voice commands\cite{Wei1997}, was used as a testbed for an autonomous instrument tracking system, where color is used to distinguish tissue from instruments and the camera is moved planarly to keep the instruments fixed in the field of view\cite{Lee1994}.  Similar to the AESOP system, the ViKY Robotic Endoscope Holder can be controlled via voice command. One research group has used the ViKY system to track the surgical instruments\cite{Voros2010}.

Another method of automation involves tracking instruments or the surgeon and moving the robotic instruments accordingly. This is useful when the movement of the tracked targets originates from a human, thereby making the autonomous movement a direct extension of the surgeon's skill.
An alternative to instrument tracking for inferring the surgeon's intent is to track the surgeon's gaze\cite{gras2016}. Eye tracking technology may be used to determine where the surgeon is looking, which may be used to autonomously reposition the laparoscopic camera\cite{Ali2007} or an auxiliary surgical instrument\cite{Noonan2008}. The eyetracking methods can be built into surgeon consoles as they teleoperate a robot, making the processing of tracking gaze seamless to the surgeon, and are currently in development within the dVRK community.

\subsection{Learning from Humans}
Learning from human-provided examples provides another approach to automating surgical tasks. Demonstration-based learning is an intuitive method of learning and ideally circumvents the time-consuming process of manually programming motion plans\cite{Schaal1997}, which is an even greater benefit for automating tasks as complicated or involved as surgical procedures. When recording expert examples, important data to record include not only the surgical tool motions and video but also a log of events inside and outside the operative field and the surgeon's posture and speech\cite{Gao2014}. Common methods to model human movements or actions are hidden Markov models or neural networks\cite{Kaiser1996}.

Artificial intelligence has not yet reached a point where an autonomous system can simply be provided exemplary surgical videos or motion sequences and learn any arbitrary complex and multi-layered task. To facilitate the analysis of surgical tasks based on human examples, complicated surgical tasks may be decomposed into more primitive motions or subtasks. Segmenting robotic surgical motions has been performed using video and motion recordings from a da Vinci system\cite{Lin2005}. Deep learning techniques have also been used for segmentation, where pre-trained architectures were trained on non-surgical image libraries\cite{Murali2016}. Segmentation of surgical tasks is still an open problem, as it is concerned with the difficult problem of assigning labels to highly variable time-series data. A surgical activity dataset by Johns Hopkins University and Intuitive Surgical Inc. consisting of motion and video data is available for researchers interested in this problem\cite{Gao2014}.

\begin{figure}[t]
\centerline{
  \subfigure[]
     {\includegraphics[height=1.6in]{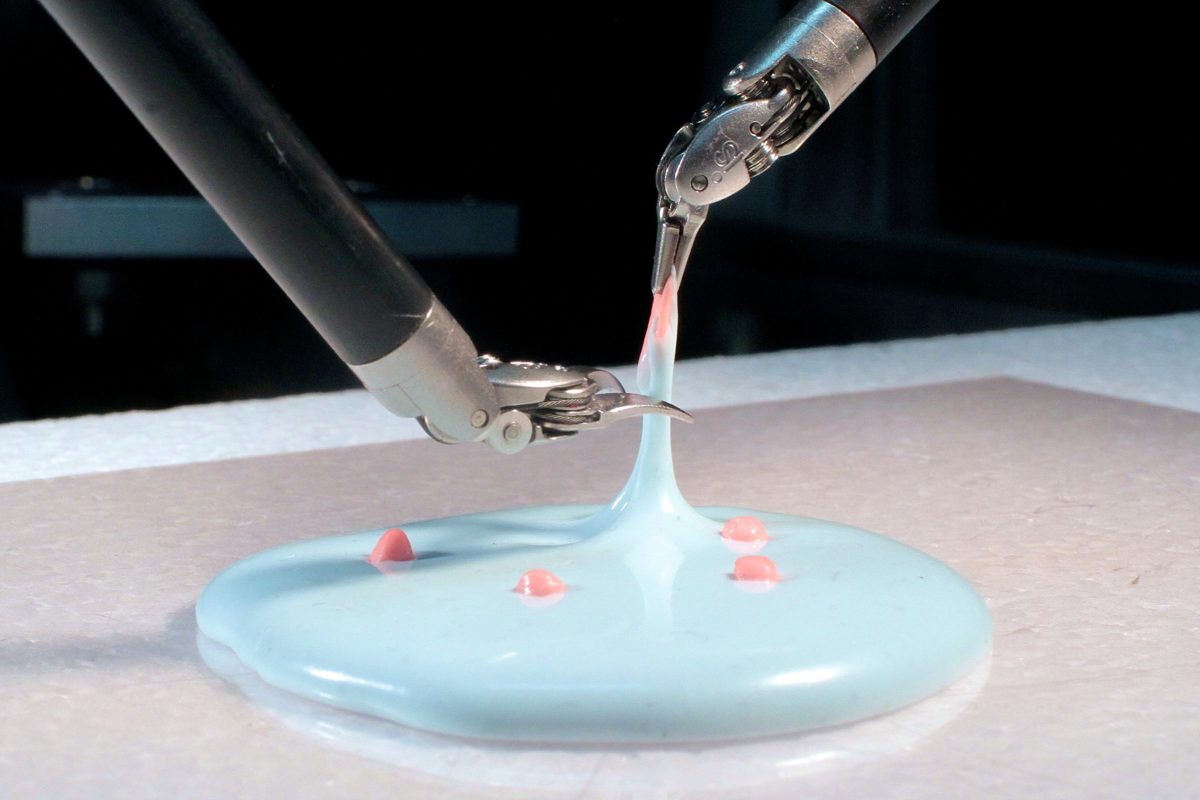}\label{debridement}}
  \hspace*{4pt}
  \subfigure[]
     {\includegraphics[height=1.6in]{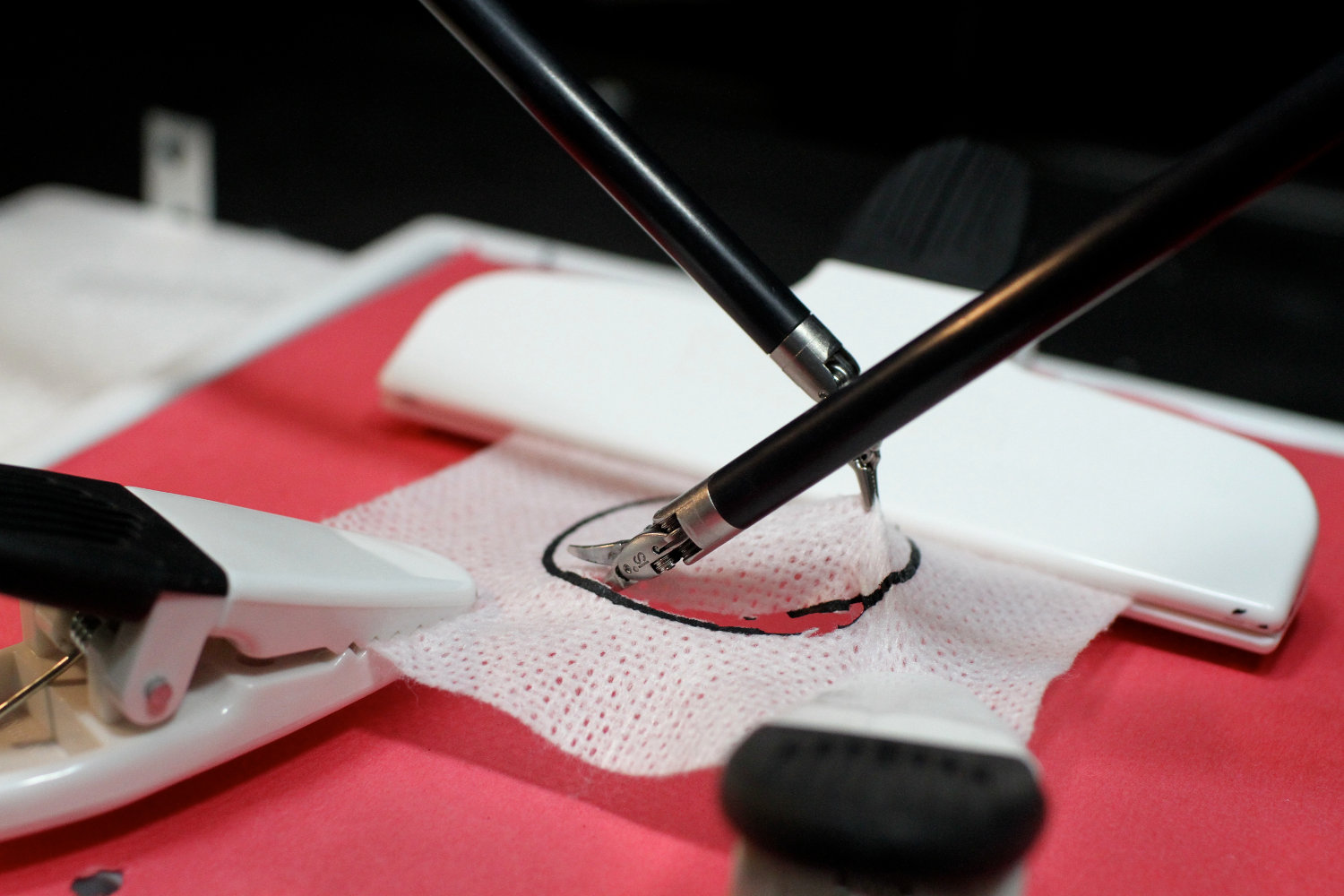}\label{patternCutting}}
}
\caption{(a) Automated debridement and (b) pattern cutting tasks using the dVRK. Reproduced by permission of Dr. Ken Goldberg, UC Berkeley.}\label{lbo}
\end{figure}

Some autonomous systems have successfully performed isolated surgical tasks based on a human-provided exemplary dataset. Trajectory smoothing of human-provided motion examples enabled faster and smoother trajectory executions on suture knot-tying tasks on an in-house laparoscopic workstation\cite{Cavu1999} compared to a human\cite{Berg2010}. Faster trajectories were achieved by iteratively updating the parameters of a controller function based on the error of a target trajectory derived from the human-provided examples\cite{Berg2010}. The EndoPAR system, developed at the Technical University of Munich as a ceiling-mounted experimental surgical platform, autonomously performed knot-tying tasks using recurrent neural networks using a database of 25 expert trajectories\cite{Mayer2006}. Recurrent neural networks are a class of models that can approximate dynamical systems, such as the set of motions involved in suture knot-tying\cite{Mayer2006}. Learning by observation (LBO) techniques were employed to automate multilateral subtasks including debridement and pattern cutting\cite{Murali2015} as shown in Fig. \ref{lbo}. To achieve autonomous debridement and pattern cutting, the LBO approach first involved segmenting human-provided motion examples into fundamental motions, such as penetration, grasping, and retraction, from which a finite state machine (FSM) is defined. An FSM is an abstract model composed of the steps involved in the task at hand. The parameters of the FSM are adjusted based on repeated executions of the motion sequences represented by the FSM\cite{Murali2015}. Finally, autonomous microanastamosis was demonstrated using learning from demonstration techniques on the KUKA LWR platform\cite{Shademan2016}.

\begin{figure}[t]
\centerline{
  \subfigure[]
     {\includegraphics[height=1.7in]{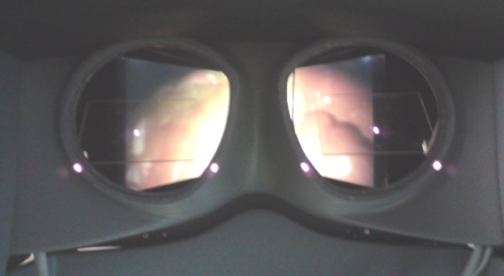}\label{perceptualDocking1}}
  \hspace*{4pt}
  \subfigure[]
     {\includegraphics[height=1.7in]{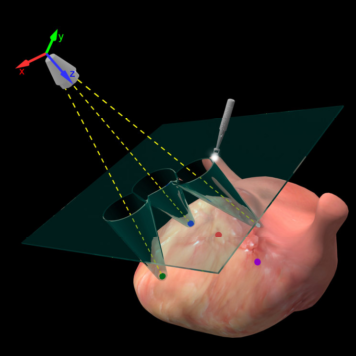}\label{perceptualDocking2}}
}
\caption{(a) Eye tracking system integrated into da Vinci Surgical System's surgical console. (b) Model of safety boundaries to constrain motion of surgical tools. Conical pathways are positioned depending on the the surgeon's fixation point.\cite{Mylonas2008}}\label{perceptualDocking}
\end{figure}

While instrument motion and video recordings may be the most direct modality to represent human-performed surgical examples, other sensor modalities may provide context-specific information based on human behavior during surgical tasks. The concept of perceptual docking, developed at Imperial College London, enables robotic systems to learn behaviors of a human based on eye tracking (Fig. \ref{perceptualDocking1}), such as visual search and information seeking behaviors\cite{Yang2002}. Additionally, the fixation point of the surgeon can be used to update a model of safety boundaries in the operative field without any prior knowledge of the anatomy\cite{Mylonas2008}. Fig. \ref{perceptualDocking2} shows an example safety boundary and conical pathways whose positions depend on the surgeon's fixation point.

\section{Challenges}
Despite success in automation of complicated but isolated surgical tasks in a research setting, bringing these automation techniques to the operating room is not straightforward as real-world challenges are often ignored in controlled laboratory experiments. This section outlines some of the many challenges associated with automating surgical tasks and some methods that are used to compensate for these difficulties.

\subsection{Uncertainties with Environment}
Involuntary movement of the patient make the operative field a changing environment. This is a challenge for automation because vital structures may be at risk when instruments are in close proximity. Preoperative plans may become obsolete when the patient moves. One approach to respond to this issue is to mechanically fix the patient to restrict intraoperative movement as is done with the ROBODOC system. A disadvantage of this technique is that the pins or frames to fix the patient in place are often invasive\cite{Taylor1994} and require an additional preoperative step\cite{Spencer1996}. 

An alternative approach is to accommodate for the changing environment. Accommodation requires the autonomous systems to either track movements or be flexible to unknown environments. The CyberKnife system is capable of adapting to patient movement using respiratory tracking using fiducial markers\cite{Moustris2011}. Compensating for smaller-scale involuntary movement can provide a more stable image, such as compensation for a beating heart in intracardiac surgery. Heartbeat compensation may be performed using a high-speed camera to track points of reference, using a synchronization algorithm to offset the camera output to make the points of reference appear stationary to the surgeon or to move robotic manipulators in synchrony with the beats\cite{Nakamura2001,Ginhoux2004,ruszkowski2016}.

Uncertainties may also arise when working with vision systems. Visual obstacles such as organs, instruments, and electrocautery smoke may occlude areas of interest\cite{Groeger2008} in endoscopic cameras that provide difficulty for vision-based systems and surgeons that rely on direct vision. Alternative real-time vision modalities may thus be useful, such as ultrasound imaging, which for instance may be used for intracardiac surgery when blood prevents direct vision\cite{Cannon2003}; however, even then there is masking and shadowing effects from non-tissue (air, or instrument) materials, as well as generally poor image resolutions. MR and CT guidance provides other challenges, including visual artifacts from devices, and insufficient imaging bandwidth to capture breathing and heartbeat in real-time. These uncertainties will modify the treatment approach and the treatment trajectories themselves.

\subsection{Uncertainties with Device}
Force sensing capabilities will be beneficial to autonomous surgical systems for autonomously identifying if a collision between a tool and a tissue (or tool) has occurred off-screen, which can be particularly dangerous as the teleoperating surgeon has no line-of-sight or haptic feedback\cite{Sung2001}. However, even for current teleoperated surgical systems such as the da Vinci system, surgeons do not receive full haptic force feedback\cite{Lanfranco2003}. Part of the challenge in supplying full haptic feedback is outfitting the surgical instruments with appropriate force feedback sensors\cite{Okamura2009}, but this requires careful design of the sensors or tools\cite{Mohareri2014a}. Integration of appropriate force sensors in manipulators is an active area of research, with approaches including soft and deformable grippers\cite{Homberg2015} and miniature uniaxial force sensors\cite{Yip2010}. Forces may also be estimated without direct force sensors based on the difference between true and desired manipulator positions, but the precision of such methods may not be high enough for delicate work\cite{Okamura2009}. 

In addition to sensing insufficiencies, modeling and controlling robots with complex configurations may be difficult. A promising technology for robotic surgery is continuum manipulators, structures that can bend continuously along their length\cite{Robinson1999} producing motion similar to that of an elephant's trunk. Continuum manipulators provide increased manipulability, which is advantageous for navigating through narrow or complicated passages. Surgical applications for continuum manipulators include endoscopes, colonoscopes\cite{Chen2009}, and catheters\cite{Bailly2005}, and most surgical methods in the past two decades have moved increasingly towards a minimally invasive procedure using continuum devices. A disadvantage of continuum manipulators is they are difficult to model and control. Recent research shows a promising implementation of a control scheme for a continuum manipulator for cardiac ablation tasks, even without a completely known model for the manipulator\cite{Yip2017}.

\subsection{Uncertainties with Procedure}
The easiest tasks to automate are those that may be routinized. One current obstacle for autonomy in surgical planning is there are generally no well-defined rules stating the optimal corrections to perform on a human or allowances for certain surgical tasks. For example, there is no consensus among expert surgeons on the range of acceptable knee-to-hip angles in knee replacement surgery\cite{RodriguezyBaena2010}. Thus, in the interest of routinizing surgical tasks, task-specific specifications or constraints must be agreed upon for the autonomous system to satisfy.

Randomized clinical trials is a necessity to reduce uncertainty with surgical techniques and protocols\cite{Rosenberg2012}. Clinical trials have been used to reject new and promising surgical techniques and to determine safe protocols, such as appropriate safety margins for melanoma excision\cite{Hall1996,Hauschild2003}. Following a methodological approach in accepting surgical techniques and defining protocols should develop a set of standardizations in surgery and thus potentially lead to a routinization of certain procedures.

\section{Ethics and Legality}
Beyond the technical challenges, ethical and legal issues may provide another obstacle in bringing automation to the operating room. As current practices of automation in surgery utilize robotics more as a tool than as an entirely independent agent, the ethical and legal issues described in this section will become increasingly more significant as systems increase in their level of autonomy.

\subsection{Ensuring Patient Safety}
Classically, physicians take the Hippocratic Oath, to which the phrase "do no harm" is often mistakenly attributed\cite{Markel2004}. Nevertheless, "do no harm" roughly summarizes the message the oath intends to deliver. The Declaration of Geneva more explicitly binds physicians to the promotion of patient health with the words "The health of my patient will be my first consideration"\cite{WHO2001}. To ensure that autonomous surgical systems satisfy the duty to prioritize the health of the patient, there must be guarantees of safety in place.

As autonomy in surgical robotics is still in infancy, human surgeons still play a large role in ensuring the safety of the patient. Human experts are still required for generating and verifying preoperative plans. The ROBODOC system can execute plans autonomously, but the plans are generated, optimized, and supervised by human surgeons\cite{Howe1999}. Similarly, the CyberKnife system is capable of processing preoperative CT images and generating motion plans to reach all target sites, but a human must first review and edit the plans prior to execution\cite{B2004}. Thus, with human verification of surgical plans, autonomous surgical systems may vicariously satisfy the requirement to prioritize patient health.

Robotic systems may also be mechanically or programmatically designed to intrinsically incorporate safety measures, such as the RIO arm which is designed such that failure of any component does not cause an unsafe environment\cite{Hagag2011}. Motion constraints can be employed to ensure surgical safety by setting active assists, such as no-fly zones, to prohibit dangerous or harmful movements\cite{Taylor2006}. Examples in practice include the ACROBOT which restricts movements to a predefined volume\cite{RodriguezyBaena2010}, the Precision Freehand Sculptor which retracts its rotary blade near prohibited regions\cite{Brisson2004}, and the da Vinci Surgical System's manipulators whose remote center of motion prevents harm at the trocar point.

\subsection{Culpability}
Four commonly utilized guiding principles for medical ethics are
\begin{enumerate}
\item{Respect for Patient Autonomy,}
\item{Justice,}
\item{Nonmaleficence, and}
\item{Beneficence\cite{Beauchamp2001}.}
\end{enumerate}
The first two guiding principles ensure that the treatments are well-suited to the patient's specific situation and preferences and that medical resources are distributed in a just manner. The last two principles are to not intentionally harm the patient and to take actions that promote patient welfare\cite{Anderson2006}. While the Principles for Patient Autonomy and Justice are more concerned with patient and hospital management, the Principles of Nonmaleficence and Beneficence are the principles more related to intraoperative surgical autonomy as they summarize the primary constraints for automating surgery: to take actions that improve the health of the patient without harming the patient. 

Enforcing the principles of Nonmaleficence and Beneficence is not straightforward because autonomy in robotics yields concerns regarding culpability in the event of errors, such as collisions in autonomous vehicles\cite{Hevelke2015}. Ambiguity arises when assigning blame of the cause of the error. The potential assignees include the designers of the system, the user of the system, or the system itself. 

Autonomy for surgical robotics faces this concern in the event of surgical complications or controversial procedures. While current practices of autonomy in surgery utilize robotics more as an intelligent tool than as an entirely independent agent, increasing the level of autonomy of the surgical system raises the question about who is in charge of the surgical operation\cite{RodriguezyBaena2010}. In the case of surgical errors, the human surgeon who used or authorized the use of the robot, the hospital, the robot designer, and the insurer are included as potential culpable entities\cite{Bekey2012}. Additionally, euthanasia and abortion are controversial procedures for clinicians as they may be considered to conflict with the clinician's effort to promote patient health. If an autonomous robot performs these procedures, a question that remains is whether the clinician still satisfies the ethical guiding principles.

Long distance telemedicine or telesurgery, while currently not common, further complicates the issue of culpability when surgical errors occur when the surgical staff and patient are in different jurisdictions\cite{Sharkey2013,Dickens2006}, such as if a surgery is performed in an underserved region and is overseen by a surgeon in another part of the world. Not only may there be a conflict of law or socially acceptable practices among different regions, but the parties involved in the litigation may attempt to select the court system most advantageous to their own agendas\cite{Dickens2006}. Finally, the issue of appropriate licenses to operate for clinicians teleoperating into remote areas brings into question legal concerns. To avoid these issues, clear delineation of accountability must be established prior to any operations that cross borders\cite{Sharkey2013}.

\subsection{Approval for Use}
\label{approvalForUse}
A surgical system must have approval from the Food and Drug Administration (FDA) prior to marketing in the US. Surgical robotic systems typically fall under the Class II category for FDA regulation, where Class I includes devices with minimal potential harm and Class III includes riskier or life-sustaining devices. To market a new surgical robotic device, a 510(k) premarket notification must be submitted to the FDA to illustrate how the new device is substantially equivalent to a predicate, or legally marketed, device.

Introducing completely novel automation techniques into the market is more difficult without an equivalent predicate system. If no equivalent predicate device is found, the system is marked as Class III and a premarket approval application must be submitted, which is a more involved process. Acquiring 510(k) approval is more feasible if small steps toward automation are taken. For instance, while the CyberKnife system was the first to introduce frameless stereotactic radiosurgery to treat tumors anywhere on the body, Accuray initially established substantial equivalence to a predicate system called Varian Clinac 600SR. Subsequent 510(k) notifications for later versions of the CyberKnife referred to preceding CyberKnife versions to introduce features such as respiratory tracking. 

\section{Conclusion and Future Directions}
As techniques in automation improve in the field of robotics, so will robot autonomy for surgery. While systems such as the ROBODOC and CyberKnife demonstrate supervised execution of surgical plans in practice, perfecting and implementing the more complex and involved tasks shown in research settings is the next step in enhancing autonomy in surgery. Introducing the more complex tasks into the clinical setting is a joint effort, involving engineering teams to design techniques and systems for automation, clinicians to guide and verify the design of new systems, and industries to bring new techniques and systems into practice. These participants of the efforts toward surgical autonomy must work together to overcome the challenges mentioned in this chapter associated with robotics, the operating environment, and marketing new systems.

In the pursuit of automation in surgery, we must exercise caution that the objective is not to remove the human from the surgical team but to (1) enhance the efficacy of a surgery, and (2) generate new surgical approaches. Furthermore, while dedicating minimal level of research to preoperative planning for complicated procedures precludes hope for full autonomy, improving automation of intraoperative tasks is a step toward introducing more supervised autonomy in the operating room. Supervised autonomy may be the most promising level of autonomy for surgery as it takes advantage of mechanical systems' skill at executing precise motions and of human surgeons' abilities to provide high-level supervisory and interventional support.

\bibliographystyle{ws-rv-van}
\bibliography{ws-rv-sample}

\begin{thebibliography}{109}
\providecommand{\natexlab}[1]{#1}
\providecommand{\url}[1]{\texttt{#1}}
\expandafter\ifx\csname urlstyle\endcsname\relax
  \providecommand{\doi}[1]{doi: #1}\else
  \providecommand{\doi}{doi: \begingroup \urlstyle{rm}\Url}\fi

\bibitem{Talamini2002}
M.~Talamini, K.~Campbell, and C.~Stanfield, {Robotic Gastrointestinal Surgery:
  Early Experience and System Description}, \emph{Journal of Laparoendoscopic
  {\&} Advanced Surgical Techniques}. {\bf 12}\penalty0 (4),  (2002).

\bibitem{Moustris2011}
G.~Moustris, S.~Hiridis, K.~Deliparaschos, and K.~Konstantinidis, {Evolution of
  autonomous and semi-autonomous robotic surgical systems : a review of the
  literature}, \emph{The International Journal of Medical Robotics and Computer
  Assisted Surgery}. {\bf 7}\penalty0 (4), \penalty0 375--392,  (2011).
\newblock \doi{10.1002/rcs}.

\bibitem{Kang2001}
H.~Kang and J.~Wen.
\newblock {EndoBot : a Robotic Assistant in Minimally Invasive Surgeries}.
\newblock In \emph{International Conference on Robotics and Automation}, pp.
  2031--2036,  (2001).
\newblock ISBN 0780364759.

\bibitem{Taylor1999}
R.~Taylor, P.~Jensen, L.~Whitcomb, A.~Barnes, D.~Stoianovici, Z.~Wang, and
  L.~Kavoussi, {A Steady-Hand Robotic System for Microsurgical Augmentation},
  \emph{The International Journal of Robotics Research}. {\bf 18}\penalty0
  (12), \penalty0 1201--1210,  (1999).

\bibitem{Kapoor2003}
A.~Kapoor, R.~Kumar, and R.~Taylor.
\newblock {Simple Biomanipulation Tasks with "Steady Hand" Cooperative
  Manipulator}.
\newblock In \emph{Medical Image Computing and Computer Assisted Intervention
  Conference}, pp. 141--148,  (2003).
\newblock ISBN 3540204628.

\bibitem{Tombropoulos2011}
R.~Z. Tombropoulos, J.~R. Adler, and J.-c. Latombe, {CARABEAMER: A treatment
  planner for a robotic radiosurgical system with general kinematics},
  \emph{Medical Image Analysis}. {\bf 3}\penalty0 (3), \penalty0 237--264,
  (2011).

\bibitem{Sharkey2013}
N.~Sharkey and A.~Sharkey, {Robotic Surgery: On the Cutting Edge of Ethics},
  \emph{Computer}. {\bf 46}\penalty0 (1), \penalty0 56--64,  (2013).

\bibitem{Gao2014}
Y.~Gao, S.~S. Vedula, C.~E. Reiley, N.~Ahmidi, B.~Varadarajan, H.~C. Lin,
  L.~Tao, L.~Zappella, B.~Bejar, D.~D. Yuh, C.~C.~G. Chen, R.~Vidal,
  S.~Khudanpur, and G.~D. Hager.
\newblock {JHU-ISI Gesture and Skill Assessment Working Set ( JIGSAWS ): A
  Surgical Activity Dataset for Human Motion Modeling}.
\newblock In \emph{Workshop on Modeling and Monitoring of Computer Assisted
  Interventions}, pp. 1--10,  (2014).

\bibitem{Siciliano2008}
R.~H. Taylor, A.~Menciassi, G.~Fichtinger, and P.~Dario.
\newblock {Medical Robotics and Computer-Integrated Surgery}.
\newblock In \emph{Springer Handbook of Robotics}, pp. 1199--1222. Springer,
  1st edition,  (2008).

\bibitem{Heemskerk2006}
J.~Heemskerk, R.~Zandbergen, J.~G. Maessen, J.~W.~M. Greve, and N.~D. Bouvy,
  {Advantages of advanced laparoscopic systems}, \emph{Surgical Endoscopy}.
  {\bf 20}, \penalty0 730--733,  (2006).
\newblock \doi{10.1007/s00464-005-0456-3}.

\bibitem{Kang2001a}
H.~Kang and J.~Wen.
\newblock {Robotic Assistants Aid Surgeons During Minimally Invasive
  Procedures}.
\newblock In \emph{IEEE Engineering in Medicine and Biology Conference}, number
  February, pp. 94--104,  (2001).

\bibitem{Gomes2011}
P.~Gomes, {Robotics and Computer-Integrated Manufacturing Surgical robotics :
  Reviewing the past , analysing the present , imagining the future},
  \emph{Robotics and Computer Integrated Manufacturing}. {\bf 27}\penalty0 (2),
  \penalty0 261--266,  (2011).
\newblock ISSN 0736-5845.
\newblock \doi{10.1016/j.rcim.2010.06.009}.
\newblock URL \url{http://dx.doi.org/10.1016/j.rcim.2010.06.009}.

\bibitem{Chun2008}
K.~R.~J. Chun, B.~Schmidt, B.~K{\"{o}}kt{\"{u}}rk, R.~Tilz, A.~F{\"{u}}rnkranz,
  M.~Konstantinidou, E.~Wissner, A.~Metzner, F.~Ouyang, and K.-h. Kuck,
  {Catheter Ablation – New Developments in Robotics}, \emph{Herz}. {\bf
  33}\penalty0 (8), \penalty0 586--589,  (2008).
\newblock \doi{10.1007/s00059-008-3180-7}.

\bibitem{Davies2000}
B.~Davies.
\newblock {A review of robotics in surgery}.
\newblock In \emph{Proceedings of the Institution of Mechanical Engineers},
  vol. 214, pp. 129--140,  (2000).

\bibitem{boogerd2016}
L.~F. Boogerd, H.~J. Handgraaf, C.~Boonstra, A.~L. Vahrmeijer, and C.~J. Van
  De~Velde, Image-guided surgery, \emph{Surgical Oncology: Theory and
  Multidisciplinary Practice}.  (2016).

\bibitem{Muntener2006}
M.~Muntener, A.~Patriciu, D.~Petrisor, D.~Mazilu, H.~Bagga, L.~Kavoussi,
  K.~Cleary, and D.~Stoianovici, {Magnetic Resonance Imaging Compatible Robotic
  System for Fully Automated Brachytherapy Seed Placement}, \emph{Urology}.
  {\bf 68}\penalty0 (6), \penalty0 1313--1317,  (2006).
\newblock \doi{10.1016/j.urology.2006.08.1089}.

\bibitem{Alterovitz2008}
R.~Alterovitz, M.~Branicky, and K.~Goldberg, {Uncertainty for Image-guided
  Medical Needle Steering}, \emph{The International Journal of Robotics
  Research}. {\bf 27}\penalty0 (11), \penalty0 1361--1374,  (2008).
\newblock \doi{10.1177/0278364908097661}.

\bibitem{Vardo2017}
J.~Vargo, J.~Townsend, S.~Sullivan, M.~Detamore, and B.~Andrews, {Modern
  Applications of Computer Bioengineering in Maxillofacial Surgery: Image
  Guided Surgical Navigation and CAD/CAM Custom Implants}, \emph{Journal of
  Computer Engineering and Information Technology}. {\bf 6}, \penalty0 2--5,
  (2017).
\newblock \doi{10.4172/2324-9307.1000164}.

\bibitem{Abayazid2014}
M.~Abayazid, G.~J. Vrooijink, S.~Patil, R.~Alterovitz, and S.~Misra,
  {Experimental evaluation of ultrasound-guided 3D needle steering in
  biological tissue}, \emph{International Journal of Computer Assisted
  Radiology}. {\bf 9}, \penalty0 931--939,  (2014).
\newblock \doi{10.1007/s11548-014-0987-y}.

\bibitem{Link2006}
R.~E. Link, S.~B. Bhayani, and L.~R. Kavoussi, {A Prospective Comparison of
  Robotic and Laparoscopic Pyeloplasty}, \emph{Annals of Surgery}. {\bf
  243}\penalty0 (4), \penalty0 486--491,  (2006).
\newblock \doi{10.1097/01.sla.0000205626.71982.32}.

\bibitem{Lanfranco2003}
A.~R. Lanfranco, A.~E. Castellanos, J.~P. Desai, and W.~C. Meyers, {Robotic
  Surgery - A Current Perspective}, \emph{Annals of Surgery}. {\bf
  239}\penalty0 (1), \penalty0 14--21,  (2003).
\newblock \doi{10.1097/01.sla.0000103020.19595.7d}.

\bibitem{Ballantyne2002}
G.~H. Ballantyne, {Robotic surgery, telerobotic surgery, telepresence, and
  telementoring: Review of early clinical results}, \emph{Surgical Endoscopy}.
  {\bf 16}, \penalty0 1389--1402,  (2002).
\newblock \doi{10.1007/s00464-001-8283-7}.

\bibitem{Manzey2009}
D.~Manzey, G.~Strauss, C.~Trantakis, T.~Lueth, S.~Rottger, J.~Bahner-Heyne,
  A.~Dietz, and J.~Meixensberger, {Automation in Surgery: A Systematic
  Approach}, \emph{Surgical Technology International}. {\bf 18},  (2009).

\bibitem{Eggers2006}
G.~Eggers and J.~Mu, {Image-to-patient registration techniques in head
  surgery}, \emph{International Journal of Oral and Maxillofacial Surgery}.
  {\bf 35}, \penalty0 1081--1095,  (2006).
\newblock \doi{10.1016/j.ijom.2006.09.015}.

\bibitem{Howe1999}
R.~D. Howe and Y.~Matsuoka, {Robotics for Surgery}, \emph{Annual Review of
  Biomedical Engineering}. {\bf 1}, \penalty0 211--240,  (1999).

\bibitem{Netravali2016}
N.~A. Netravali, M.~B{\"{o}}rner, and W.~L. Bargar.
\newblock {The Use of ROBODOC in Total Hip and Knee Arthroplasty}.
\newblock In \emph{Computer-Assisted Musculoskeletal Surgery}, pp. 219--234.
  (2016).
\newblock ISBN 9783319129433.
\newblock \doi{10.1007/978-3-319-12943-3}.

\bibitem{Spencer1996}
E.~H. Spencer, {The ROBODOC Clinical Trial: A Robotic Assistant for Total Hip
  Arthroplasty}, \emph{Orthopaedic Nursing1}. {\bf 15}\penalty0 (1), \penalty0
  9--14,  (1996).

\bibitem{Kazanzides2008}
P.~Kazanzides, G.~Fichtinger, G.~D. Hager, A.~M. Okamura, L.~L. Whitcomb, and
  R.~H. Taylor, {Surgical and Interventional Robotics: Core Concepts,
  Technology, and Design.}, \emph{IEEE robotics {\&} automation magazine / IEEE
  Robotics {\&} Automation Society}. {\bf 15}\penalty0 (2), \penalty0 122--130,
   (2008).
\newblock ISSN 1070-9932.
\newblock \doi{10.1109/MRA.2008.926390}.

\bibitem{VanderList2016}
J.~P. van~der List, H.~Chawla, and A.~D. Pearle, {Robotic-Assisted Knee
  Arthroplasty: An Overview}, \emph{The American Journal of Orthopedics}.
  \penalty0 (June),  (2016).

\bibitem{Hagag2011}
B.~Hagag, R.~Abovitz, H.~Kang, B.~Schmitz, and M.~Conditt.
\newblock {RIO: Robotic-Arm Interactive Orthopedic System MAKOplasty : User
  Interactive Haptic Orthopedic Robotics}.
\newblock In \emph{Surgical Robotics: Systems Applications and Visions}, pp.
  219--246.  (2011).
\newblock ISBN 9781441911261.
\newblock \doi{10.1007/978-1-4419-1126-1}.

\bibitem{Rosenberg1993}
L.~B. Rosenberg.
\newblock {Virtual Fixtures: Perceptual Tools for Telerobotic Manipulation}.
\newblock In \emph{Proceedings of the IEEE Virtual Reality Annual Symposium},
  pp. 76--82,  (1993).
\newblock ISBN 0780313631.

\bibitem{Barnett2007}
G.~Barnett, M.~Linskey, J.~Adler, J.~Cozzens, W.~Friedman, M.~Heilburn,
  L.~Lunsford, M.~Schulder, and A.~Sloan, {Stereotactic radiosurgery - an
  organized neurosurgery- sanctioned definition}, \emph{Journal of
  Neurosurgery}. {\bf 106}, \penalty0 1--5,  (2007).

\bibitem{Li2002}
Q.~H. Li, L.~Zamorano, A.~Pandya, R.~Perez, J.~Gong, and F.~Diaz, {The
  Application Accuracy of the NeuroMate Robot — A Quantitative Comparison
  with Frameless and Frame-Based Surgical Localization Systems}, \emph{Computer
  Aided Surgery}. {\bf 7}, \penalty0 90--98,  (2002).
\newblock \doi{10.1002/igs.10035}.

\bibitem{Varma2006}
T.~Varma and P.~Eldridge, {Use of the NeuroMate stereotactic robot in a
  frameless mode for functional neurosurgery}, \emph{The International Journal
  of Medical Robotics and Computer Assisted Surgery}. {\bf 2}, \penalty0
  107--113,  (2006).
\newblock \doi{10.1002/rcs}.

\bibitem{Dieterich2011}
S.~Dieterich and I.~C. Gibbs, {The CyberKnife in Clinical Use: Current Roles,
  Future Expectations}, \emph{Frontiers of Radiation Therapy and Oncology}.
  {\bf 43}, \penalty0 181--194,  (2011).

\bibitem{Lindquist1995}
C.~Lindquist, {Gamma Knife Radiosurgery}, \emph{Seminars in Radiation
  Oncology}. {\bf 5}, \penalty0 197--202,  (1995).

\bibitem{Sayeh2007}
S.~Sayeh, J.~Wang, W.~T. Main, W.~Kilby, and C.~R. Maurer.
\newblock {Respiratory Motion Tracking for Robotic Radiosurgery}.
\newblock In \emph{Treating Tumors that Move with Respiration}, pp. 15--29.
  (2007).

\bibitem{B2004}
D.~Camarillo, T.~Krummel, and J.~Salisbury, {Robotic technology in surgery:
  past, present, and future}, \emph{The American Journal of Surgery}. {\bf
  188}, \penalty0 2--15,  (2004).
\newblock \doi{10.1016/j.amjsurg.2004.08.025}.

\bibitem{Hungr2009}
N.~Hungr, J.~Troccaz, N.~Zemiti, and N.~Tripodi.
\newblock {Design of an Ultrasound-Guided Robotic Brachytherapy Needle-
  Insertion System}.
\newblock In \emph{IEEE Engineering in Medicine and Biology Conference}, pp.
  250--253,  (2009).
\newblock ISBN 9781424432967.

\bibitem{Cunha2010}
J.~A. Cunha, I.-c. Hsu, J.~Pouliot, M.~{Roach III}, K.~Shinohara,
  J.~Kurhanewicz, G.~Reed, and D.~Stoianovici, {Toward adaptive stereotactic
  robotic brachytherapy for prostate cancer: Demonstration of an adaptive
  workflow incorporating inverse planning and an MR stealth robot},
  \emph{Minimally Invasive Therapy}. {\bf 19}, \penalty0 189--202,  (2010).
\newblock \doi{10.3109/13645706.2010.497000}.

\bibitem{Stoianovici2011}
D.~Stoianovici, D.~Song, D.~Petrisor, D.~Ursu, D.~Mazilu, M.~Mutener, M.~Schar,
  and A.~Patriciu.
\newblock {"MRI Stealth" robot for prostate interventions}.
\newblock In \emph{Minimally Invasive Therapy {\&} Allied Technologies}, pp.
  241--248,  (2011).
\newblock \doi{10.1080/13645700701520735.}

\bibitem{Haidegger2011}
T.~Haidegger, J.~Sandor, and Z.~Benyo, {Surgery in space: the future of robotic
  telesurgery}, \emph{Surgical Endoscopy}. {\bf 25}, \penalty0 681--690,
  (2011).
\newblock \doi{10.1007/s00464-010-1243-3}.

\bibitem{Romaniuk2017}
S.~Romaniuk and F.~Grice, \emph{{The Future of US Warfare}}. (Taylor and
  Francis, 2017).

\bibitem{Garcia2009}
P.~Garcia, J.~Rosen, M.~Noakes, G.~Elbert, M.~Treat, T.~Ganous, M.~Hanson, and
  J.~Manak, {Trauma Pod : a semi-automated telerobotic surgical system},
  \emph{The International Journal of Medical Robotics and Computer Assisted
  Surgery}. {\bf 5}, \penalty0 136--146,  (2009).
\newblock \doi{10.1002/rcs}.

\bibitem{Rosen2011}
J.~Rosen, B.~Hannaford, and R.~Satava, \emph{{Surgical Robotics: Systems
  Applications and Visions}}. (Springer Science {\&} Business Media, 2011).

\bibitem{Yip2017}
M.~C. Yip, J.~A. Sganga, and D.~B. Camarillo, {Autonomous Control of Continuum
  Robot Manipulators for Complex Cardiac Ablation Tasks}, \emph{Journal of
  Medical Robotics Research}. {\bf 2}\penalty0 (1), \penalty0 1--13,  (2017).
\newblock \doi{10.1142/S2424905X17500027}.

\bibitem{loschak2013}
P.~M. Loschak, L.~J. Brattain, and R.~D. Howe.
\newblock Automated pointing of cardiac imaging catheters.
\newblock In \emph{Robotics and Automation (ICRA), 2013 IEEE International
  Conference on}, pp. 5794--5799. IEEE,  (2013).

\bibitem{jayender2008}
J.~Jayender, M.~Azizian, and R.~V. Patel, Autonomous image-guided
  robot-assisted active catheter insertion, \emph{IEEE Transactions on
  Robotics}. {\bf 24}\penalty0 (4), \penalty0 858--871,  (2008).

\bibitem{burgner2015}
J.~Burgner-Kahrs, D.~C. Rucker, and H.~Choset, Continuum robots for medical
  applications: A survey, \emph{IEEE Transactions on Robotics}. {\bf
  31}\penalty0 (6), \penalty0 1261--1280,  (2015).

\bibitem{Harris1997}
S.~J. Harris, F.~Arambula-Cosio, Q.~Mei, R.~D. Hibberd, B.~L. Davies, J.~E.~A.
  Wickham, M.~S. Nathan, and B.~Kundu, {The Probot — an active robot for
  prostate resection}, \emph{Proceedings of the Institution of Mechanical
  Engineers}. {\bf 211}\penalty0 (4), \penalty0 317--325,  (1997).

\bibitem{Mei1996}
Q.~Mei, S.~J. Harris, M.~S. Nathan, R.~D. Hibberd, J.~E.~A. Wickham, and
  B.~Davies.
\newblock {PROBOT - A Computer Integrated Prostatectomy System}.
\newblock In \emph{Visualization in Biomedical Computing}, pp. 581--590,
  (1996).

\bibitem{RodriguezyBaena2010}
F.~{Rodriguez y Baena} and B.~Davies, {Robotic surgery: from autonomous systems
  to intelligent tools}, \emph{Robotica}. {\bf 28}, \penalty0 163--170,
  (2010).
\newblock \doi{10.1017/S0263574709990427}.

\bibitem{Jakopec2002}
M.~Jakopec, S.~J. Harris, F.~Rodriguez, P.~Gomes, and B.~L. Davies.
\newblock {Acrobot: a "Hands-on" Robot for Total Knee Replacement Surgery}.
\newblock In \emph{Advanced Motion Control}, pp. 116--120,  (2002).
\newblock ISBN 0780374797.

\bibitem{Chen2013}
Z.~Chen, A.~Deguet, R.~Taylor, S.~DiMaio, G.~Fischer, and P.~Kazanzides, {An
  Open-Source Hardware and Software Platform for Telesurgical Robotics
  Research}, \emph{The MIDAS Journal}. pp. 1--11,  (2013).

\bibitem{Chen2017}
Z.~Chen, A.~Deguet, R.~H. Taylor, and P.~Kazanzides.
\newblock {Software Architecture of the da Vinci Research Kit}.
\newblock In \emph{IEEE International Conference on Robotic Computing}, pp.
  180--187,  (2017).
\newblock ISBN 9781509067244.
\newblock \doi{10.1109/IRC.2017.69}.

\bibitem{Kazanzides2010}
P.~Kazanzides, S.~P. Dimaio, A.~Deguet, B.~Vagvolgyi, M.~Balicki, C.~Schneider,
  R.~Kumar, A.~Jog, B.~Itkowitz, C.~Hasser, and R.~H. Taylor, {The Surgical
  Assistant Workstation (SAW) in Minimally-Invasive Surgery and Microsurgery},
  \emph{The MIDAS Journal}. pp. 1--9,  (2010).

\bibitem{Murali2015}
A.~Murali, S.~Sen, B.~Kehoe, A.~Garg, S.~Mcfarland, S.~Patil, W.~D. Boyd,
  S.~Lim, P.~Abbeel, and K.~Goldberg, {Learning by Observation for Surgical
  Subtasks: Multilateral Cutting of 3D Viscoelastic and 2D Orthotropic Tissue
  Phantoms}, \emph{International Conference on Robotics and Automation}. pp.
  1202--1209,  (2015).

\bibitem{Mohareri2014}
O.~Mohareri and S.~Salcudean.
\newblock {da Vinci Auxiliary Arm as a Robotic Surgical Assistant for
  Semi-Autonomous Ultrasound Guidance During Robot-Assisted Laparoscopic
  Surgery}.
\newblock In \emph{The Hamlyn Symposium on Medical Robotics}, pp. 45--46,
  (2014).
\newblock ISBN 9780956377654.

\bibitem{Hannaford2013}
B.~Hannaford, J.~Rosen, D.~W. Friedman, H.~King, P.~Roan, L.~Cheng, D.~Glozman,
  J.~Ma, S.~N. Kosari, and L.~White, {Raven-II : An Open Platform for Surgical
  Robotics Research}, \emph{IEEE Transactions on Biomedical Engineering}. {\bf
  60}\penalty0 (4), \penalty0 954--959,  (2013).

\bibitem{Hu2015}
D.~Hu, Y.~Gong, B.~Hannaford, and E.~J. Seibel.
\newblock {Semi-autonomous Simulated Brain Tumor Ablation with RavenII Surgical
  Robot using Behavior Tree}.
\newblock In \emph{International Conference on Robotics and Automation}, pp.
  3868--3875,  (2015).
\newblock ISBN 9781479969234.

\bibitem{Kehoe2014}
B.~Kehoe, G.~Kahn, J.~Mahler, J.~Kim, A.~Lee, A.~Lee, K.~Nakagawa, S.~Patil,
  W.~D. Boyd, P.~Abbeel, and K.~Goldberg.
\newblock {Autonomous Multilateral Debridement with the Raven Surgical Robot}.
\newblock In \emph{International Conference on Robotics and Automation}, pp.
  1432--1439,  (2014).
\newblock ISBN 9781479936854.

\bibitem{Hagn2010}
U.~Hagn, R.~Konietschke, A.~Tobergte, M.~Nickl, S.~Jorg, B.~Kubler, G.~Passig,
  M.~Groger, F.~Frohlich, U.~Seibold, L.~Le-Tien, A.~Albu-Schaffer,
  A.~Nothhelfer, F.~Hacker, M.~Grebenstein, and G.~Hirzinger, {DLR MiroSurge: a
  versatile system for research in endoscopic telesurgery}, \emph{International
  Journal of Computer Assisted Radiology}. {\bf 5}, \penalty0 183--193,
  (2010).
\newblock \doi{10.1007/s11548-009-0372-4}.

\bibitem{Shademan2016}
A.~Shademan, R.~S. Decker, J.~D. Opfermann, S.~Leonard, A.~Krieger, and P.~C.
  Kim, Supervised autonomous robotic soft tissue surgery, \emph{Science
  translational medicine}. {\bf 8}\penalty0 (337), \penalty0 337ra64--337ra64,
  (2016).

\bibitem{yip2012}
M.~C. Yip, D.~G. Lowe, S.~E. Salcudean, R.~N. Rohling, and C.~Y. Nguan, Tissue
  tracking and registration for image-guided surgery, \emph{IEEE transactions
  on medical imaging}. {\bf 31}\penalty0 (11), \penalty0 2169--2182,  (2012).

\bibitem{Vitrani2005}
M.~Vitrani, G.~Morel, and T.~Ortmaier.
\newblock {Automatic Guidance of a Surgical Instrument with Ultrasound Based
  Visual Servoing}.
\newblock In \emph{International Conference on Robotics and Automation}, pp.
  508--513,  (2005).
\newblock ISBN 078038914X.

\bibitem{Stoll2006}
J.~Stoll, P.~Novotny, R.~Howe, and P.~Dupont.
\newblock {Real-time 3D Ultrasound-based Servoing of a Surgical Instrument}.
\newblock In \emph{International Conference on Robotics and Automation}, pp.
  613--618,  (2006).
\newblock ISBN 0780395050.

\bibitem{Omote1999}
K.~Omote, H.~Feussner, A.~Ungeheuer, K.~Arbter, G.-q. Wei, J.~Siewert, and
  G.~Hirzinger, {Self-Guided Robotic Camera Control for Laparoscopic Surgery
  Compared with Human Camera Control}, \emph{The American Journal of Surgery}.
  {\bf 177}, \penalty0 321--324,  (1999).

\bibitem{Pandya2014}
A.~Pandya, L.~A. Reisner, B.~King, N.~Lucas, A.~Composto, M.~Klein, and R.~D.
  Ellis, {A Review of Camera Viewpoint Automation in Robotic and Laparoscopic
  Surgery}, \emph{Robotics}. {\bf 3}, \penalty0 310--329,  (2014).
\newblock \doi{10.3390/robotics3030310}.

\bibitem{Wei1997}
G.~Wei, K.~Arbter, and G.~Hirzinger, {Real-Time Visual Servoing for
  Laparoscopic Surgery}, \emph{IEEE Engineering in Medicine and Biology
  Conference}. {\bf 16}, \penalty0 40--45,  (1997).

\bibitem{Lee1994}
C.~Lee, Y.~Wang, D.~Uecker, and Y.~Wang.
\newblock {Image Analysis for Automated Tracking in Robot-Assisted Endoscopic
  Surgery}.
\newblock In \emph{International Conference on Pattern Recognition}, pp.
  88--92,  (1994).

\bibitem{Voros2010}
S.~Voros, G.-P. Haber, J.-F. Menudet, J.-A. Long, and P.~Cinquin, {ViKY Robotic
  Scope Holder: Initial Clinical Experience and Preliminary Results Using
  Instrument Tracking}, \emph{IEEE Transactions on Mechatronics}. {\bf
  15}\penalty0 (6), \penalty0 879--886,  (2010).

\bibitem{gras2016}
G.~Gras and G.-Z. Yang.
\newblock Intention recognition for gaze controlled robotic minimally invasive
  laser ablation.
\newblock In \emph{Intelligent Robots and Systems (IROS), 2016 IEEE/RSJ
  International Conference on}, pp. 2431--2437. IEEE,  (2016).

\bibitem{Ali2007}
S.~Ali, L.~Reisner, B.~King, A.~Cao, G.~Auner, M.~Klein, and A.~Pandya, {Eye
  gaze tracking for endoscopic camera positioning: An application of a
  hardware/software interface developed to automate AESOP}, \emph{Studies in
  Health Technology and Informatics}. {\bf 132}, \penalty0 4--7,  (2007).

\bibitem{Noonan2008}
D.~P. Noonan, G.~P. Mylonas, A.~Darzi, and G.~Yang.
\newblock {Gaze Contingent Articulated Robot Control for Robot Assisted
  Minimally Invasive Surgery}.
\newblock In \emph{International Conference on Intelligent Robotics and
  Systems}, pp. 22--26,  (2008).
\newblock ISBN 9781424420582.

\bibitem{Schaal1997}
S.~Schaal.
\newblock {Learning From Demonstration}.
\newblock In \emph{Conference on Neural Information Processing Systems}, pp.
  1040--1046,  (1997).

\bibitem{Kaiser1996}
M.~Kaiser and R.~Dillmann.
\newblock {Building Elementary Robot Skills from Human Demonstration}.
\newblock In \emph{International Conference on Robotics and Automation}, pp.
  2700--2705,  (1996).

\bibitem{Lin2005}
H.~Lin, I.~Shafran, T.~Murphy, A.~Okamura, D.~Yuh, and G.~Hager.
\newblock {Automatic Detection and Segmentation of Robot-Assisted Surgical
  Motions}.
\newblock In \emph{Medical Image Computing and Computer Assisted Intervention
  Conference}, pp. 802--810,  (2005).

\bibitem{Murali2016}
A.~Murali, A.~Garg, S.~Krishnan, F.~T. Pokorny, P.~Abbeel, T.~Darrell, and
  K.~Goldberg.
\newblock {TSC-DL: Unsupervised Trajectory Segmentation of Multi-Modal Surgical
  Demonstrations with Deep Learning}.
\newblock In \emph{International Conference on Robotics and Automation},
  (2016).

\bibitem{Cavu1999}
M.~C. Cavu, F.~Tendick, M.~Cohn, and S.~S. Sastry, {A Laparoscopic Telesurgical
  Workstation}, \emph{IEEE Transactions on Robotics and Automation}. {\bf
  15}\penalty0 (4), \penalty0 728--739,  (1999).

\bibitem{Berg2010}
J.~V.~D. Berg, S.~Miller, D.~Duckworth, H.~Hu, A.~Wan, X.-y. Fu, K.~Goldberg,
  and P.~Abbeel.
\newblock {Superhuman Performance of Surgical Tasks by Robots using Iterative
  Learning from Human-Guided Demonstrations}.
\newblock In \emph{International Conference on Robotics and Automation}, pp.
  2074--2081,  (2010).
\newblock ISBN 9781424450404.

\bibitem{Mayer2006}
H.~Mayer, F.~Gomez, D.~Wierstra, I.~Nagy, A.~Knoll, and J.~Schmidhuber.
\newblock {A System for Robotic Heart Surgery that Learns to Tie Knots Using
  Recurrent Neural Networks}.
\newblock In \emph{International Conference on Intelligent Robotics and
  Systems}, pp. 1521--1537,  (2006).

\bibitem{Yang2002}
G.~Yang, L.~Dempere-Marco, X.-p. Hu, and A.~Rowe, {Visual search:
  psychophysical models and practical applications}, \emph{Image and Vision
  Computing}. {\bf 20}\penalty0 (4), \penalty0 291--305,  (2002).

\bibitem{Mylonas2008}
G.~Mylonas, K.~Kwok, A.~Darzi, and G.~Yang.
\newblock {Gaze-Contingent Motor Channelling and Haptic Constraints for
  Minimally Invasive Robotic Surgery}.
\newblock In \emph{Medical Image Computing and Computer Assisted Intervention
  Conference}, pp. 676--683,  (2008).

\bibitem{Taylor1994}
R.~H. Taylor, B.~D. Mittelstadt, H.~A. Paul, W.~Hanson, P.~Kazanzides, J.~F.
  Zuhars, B.~Williamson, B.~L. Musits, E.~Glassman, and W.~L. Bargar, {An
  Image-Directed Robotic System for Precise Orthopaedic Surgery}, \emph{IEEE
  Transactions on Robotics and Automation}. {\bf 10}\penalty0 (3), \penalty0
  261--275,  (1994).

\bibitem{Nakamura2001}
Y.~Nakamura, K.~Kishi, and H.~Kawakami.
\newblock {Heartbeat Synchronization for Robotic Cardiac Surgery}.
\newblock In \emph{International Conference on Robotics and Automation}, pp.
  2014--2019,  (2001).
\newblock ISBN 0780364759.

\bibitem{Ginhoux2004}
R.~Ginhoux, J.~A. Gangloff, M.~E. de~Mathelin, L.~Soler, M.~M. {Arenas
  Sanchez}, and J.~Marescaux.
\newblock {Beating Heart Tracking in Robotic Surgery Using 500 Hz Visual
  Servoing, Model Predictive Control and an Adaptive Observer}.
\newblock In \emph{IEEE International Conference on Robotics and Automation},
  (2004).
\newblock ISBN 0780382323.

\bibitem{ruszkowski2016}
A.~Ruszkowski, C.~Schneider, O.~Mohareri, and S.~Salcudean.
\newblock Bimanual teleoperation with heart motion compensation on the da
  vinci{\textregistered} research kit: Implementation and preliminary
  experiments.
\newblock In \emph{Robotics and Automation (ICRA), 2016 IEEE International
  Conference on}, pp. 4101--4108. IEEE,  (2016).

\bibitem{Groeger2008}
M.~Groeger, K.~Arbter, and G.~Hirzinger.
\newblock {Motion Tracking for Minimally Invasive Robotic Surgery}.
\newblock In \emph{Medical Robotics}, pp. 117--148. I-Tech Education and
  Publishing,  (2008).
\newblock ISBN 9783902613189.

\bibitem{Cannon2003}
J.~Cannon, R.~Howe, P.~Dupont, J.~Triedman, G.~Marx, and P.~del Nido,
  {Application of Robotics in Congenital Cardiac Surgery}, \emph{Pediatric
  Cardiac Surgery Annual of the Seminars in Thoracic and Cardiovascular
  Surgery}. {\bf 6}, \penalty0 72--83,  (2003).

\bibitem{Sung2001}
G.~Sung and I.~Gill, {Robotic Laparoscopy Surgery: A Comparison of the da Vinci
  and Zeus Systems}, \emph{Urology}. {\bf 58}\penalty0 (6), \penalty0 893--898,
   (2001).

\bibitem{Okamura2009}
A.~M. Okamura, {Haptic Feedback in Robot-Assisted Minimally Invasive Surgery},
  \emph{Current Opinion in Urology}. {\bf 19}\penalty0 (1), \penalty0 102--107,
   (2009).

\bibitem{Mohareri2014a}
O.~Mohareri, C.~Schneider, and S.~Salcudean.
\newblock {Bimanual Telerobotic Surgery With Asymmetric Force Feedback: A da
  Vinci Surgical System Implementation}.
\newblock In \emph{IEEE International Conference on Intelligent Robots and
  Systems}, pp. 4272--4277,  (2014).
\newblock ISBN 9781479969340.

\bibitem{Homberg2015}
B.~S. Homberg, R.~K. Katzschmann, M.~R. Dogar, and D.~Rus.
\newblock {Haptic Identification of Objects using a Modular Soft Robotic
  Gripper}.
\newblock In \emph{International Conference on Intelligent Robotics and
  Systems}, pp. 1698--1705,  (2015).
\newblock ISBN 9781479999941.

\bibitem{Yip2010}
M.~C. Yip, S.~G. Yuen, R.~D. Howe, S.~Member, A.~F. Sensing, and
  H.~Annuloplasty, {A Robust Uniaxial Force Sensor for Minimally Invasive
  Surgery}, \emph{IEEE Transactions on Biomedical Engineering}. {\bf
  57}\penalty0 (5), \penalty0 1008--1011,  (2010).

\bibitem{Robinson1999}
G.~Robinson and J.~Davies.
\newblock {Continuum Robots - A State of the Art}.
\newblock In \emph{International Conference on Robotics and Automation}, pp.
  2849--2854,  (1999).

\bibitem{Chen2009}
G.~Chen, M.~T. Pham, and T.~Redarce, {Sensor-based guidance control of a
  continuum robot for a semi-autonomous colonoscopy}, \emph{Robotics and
  Autonomous Systems}. {\bf 57}\penalty0 (6-7), \penalty0 712--722,  (2009).
\newblock ISSN 0921-8890.
\newblock URL \url{http://dx.doi.org/10.1016/j.robot.2008.11.001}.

\bibitem{Bailly2005}
Y.~Bailly and Y.~Amirat.
\newblock {Modeling and Control of a Hybrid Continuum Active Catheter for
  Aortic Aneurysm Treatment}.
\newblock In \emph{International Conference on Robotics and Automation}, pp.
  24--29,  (2005).
\newblock ISBN 078038914X.

\bibitem{Rosenberg2012}
J.~Rosenberg, A.~Fischer, and E.~Haglind, {Current controversies in colorectal
  surgery: the way to resolve uncertainty and move forward}, \emph{Colorectal
  Disease}. {\bf 14}\penalty0 (3), \penalty0 266--269,  (2012).

\bibitem{Hall1996}
J.~C. Hall, B.~Mills, H.~Nguyen, and J.~L. Hall, {Methodologic standards in
  surgical trials}, \emph{Surgery}. {\bf 119}\penalty0 (4), \penalty0 466--472,
   (1996).

\bibitem{Hauschild2003}
A.~Hauschild, F.~Rosien, and S.~Lischner, {Surgical Standards in the Primary
  Care of Melanoma Patients}, \emph{Onkologie}. {\bf 26}, \penalty0 218--222,
  (2003).

\bibitem{Markel2004}
H.~Markel, {"I Swear by Apollo" - On Taking the Hippocratic Oath}, \emph{The
  New England Journal of Medicine}. {\bf 350}, \penalty0 2026--2029,  (2004).

\bibitem{WHO2001}
WHO, {World Medical Association Declaration of Helsinki: Ethical Principles for
  Medical Research Involving Human Subjects}, \emph{World Medical Association}.
  {\bf 79}\penalty0 (June 1964),  (2001).

\bibitem{Taylor2006}
R.~Taylor, {A Perspective on Medical Robotics}, \emph{Proceedings of the IEEE}.
  {\bf 94}\penalty0 (9), \penalty0 1--14,  (2006).

\bibitem{Brisson2004}
G.~Brisson, T.~Kanade, A.~Digioia, and B.~Jaramaz.
\newblock {Precision Freehand Sculpting of Bone}.
\newblock In \emph{Medical Image Computing and Computer Assisted Intervention
  Conference}, pp. 105--112,  (2004).

\bibitem{Beauchamp2001}
T.~Beauchamp and J.~Childress, \emph{{Principles of biomedical ethics}}.
  (Oxford University Press, 2001), 4th edition.

\bibitem{Anderson2006}
M.~Anderson, S.~L. Anderson, and C.~Armen, {MedEthEx: A Prototype Medical
  Ethics Advisor}, \emph{Proceedings of the 18th Conference on Innovative
  Applications of Artificial Intelligence}. pp. 1759--1765,  (2006).

\bibitem{Hevelke2015}
A.~Hevelke and N.~Julian, {Responsibility for Crashes of Autonomous Vehicles:
  An Ethical Analysis}, \emph{Science and Engineering Ethics}. {\bf 21},
  \penalty0 619--630,  (2015).

\bibitem{Bekey2012}
G.~A. Bekey.
\newblock {Current Trends in Robotics : Technology and Ethics}.
\newblock In \emph{Robot ethics: the ethical and social implications of
  robotics}, pp. 17--34. The MIT Press, Cambridge,  (2012).

\bibitem{Dickens2006}
B.~M. Dickens and R.~J. Cook, {Legal and ethical issues in telemedicine and
  robotics}, \emph{International Journal of Gynecology and Obstetrics}. {\bf
  94}, \penalty0 73--78,  (2006).

\end{thebibliography}
\blankpage
\printindex                         
\end{document}